\documentclass[lettersize,journal]{IEEEtran}
\usepackage{amsmath,amsfonts}
\usepackage{algorithmic}
\usepackage{algorithm}
\usepackage{array}
\usepackage[caption=false,font=normalsize,labelfont=sf,textfont=sf]{subfig}
\usepackage{textcomp}
\usepackage{stfloats}
\usepackage{url}
\usepackage{verbatim}
\usepackage{graphicx}
\usepackage{cite}
\usepackage{hyperref}
\hypersetup{hypertex=true,
	colorlinks=true,
	linkcolor=red,
	anchorcolor=red,
	citecolor=red}
\usepackage{booktabs}
\usepackage{makecell}
\usepackage{booktabs}
\usepackage{bbm}
\usepackage{multirow}

\usepackage{pifont}

\usepackage[capitalize]{cleveref}
\makeatother

\begin{document}
	
	\title{DeepMatcher: A Deep Transformer-based Network for Robust and Accurate Local Feature Matching}
	
	\author{Tao Xie$^{\dagger}$, Kun Dai$^{\dagger}$, Ke Wang, Ruifeng Li, Lijun Zhao
		
		\thanks{This work was in part by National Natural Science Foundation of China under Grant 62176072 and 62073101. (Corresponding author: Ruifeng Li and Ke Wang. ${\dagger}$: These authors contribute equally.)}
		\thanks{Tao Xie, Kun Dai, Ke Wang, Ruifeng Li, Lijun Zhao are with State Key Laboratory of Robotics and System, Harbin Institute of Technology, Harbin 150006, China (email: {xietao1997@hit.edu.cn; 20s108237@stu.hit.edu.cn; 
				wangke@hit.edu.cn; 
				lrf100@hit.edu.cn;
				zhaolj@hit.edu.cn).}}
		\thanks{Tao Xie is also with SenseTime Group Inc., Beijing 100080, China (email: xietao@sensetime.com).}}
	
	\markboth{Journal of \LaTeX\ Class Files,~Vol.~14, No.~8, August~2021}%
	{Shell \MakeLowercase{\textit{et al.}}: A Sample Article Using IEEEtran.cls for IEEE Journals}
	
	
	\maketitle
	
	\begin{abstract}
		Local feature matching between images remains a challenging task, especially in the presence of significant appearance variations, e.g., extreme viewpoint changes.
		In this work, we propose DeepMatcher, a deep Transformer-based network built upon our investigation of local feature matching in detector-free methods. 
		The key insight is that local feature matcher with deep layers can capture more human-intuitive and simpler-to-match features.
		Based on this, we propose a Slimming Transformer (SlimFormer) dedicated for DeepMatcher, which leverages vector-based attention to model relevance among all keypoints and achieves long-range context aggregation in an efficient and effective manner. 
		A relative position encoding is applied to each SlimFormer so as to explicitly disclose relative distance information, further improving the representation of keypoints. 
		A layer-scale strategy is also employed in each SlimFormer to enable the network to assimilate message exchange from the residual block adaptively, thus allowing it to simulate the human behaviour that humans can acquire different matching cues each time they scan an image pair. 
		To facilitate a better adaption of the SlimFormer, we introduce a Feature Transition Module (FTM) to ensure a smooth transition in feature scopes with different receptive fields. 
		By interleaving the self- and cross-SlimFormer multiple times, DeepMatcher can easily establish pixel-wise dense matches at coarse level. 
		Finally, we perceive the match refinement as a combination of classification and regression problems and design Fine Matches Module to predict confidence and offset concurrently, thereby generating robust and accurate matches. 
		Experimentally, we show that DeepMatcher significantly outperforms the state-of-the-art methods on several benchmarks, demonstrating the superior matching capability of DeepMatcher. 
		The code is available at https://github.com/XT-1997/DeepMatcher.

		
	\end{abstract}
	
	\begin{IEEEkeywords}
		Local feature matching, Pose Estimation, Transformer.
	\end{IEEEkeywords}
	
	\section{Introduction}
	\IEEEPARstart{L}{ocal} feature matching~\cite{chen2022csr, sun2021loftr, sarlin2020superglue, zheng2022msa} is the prerequisite for a variety of geometric computer vision applications, including Simultaneous Localization and Mapping (SLAM) \cite{mur2017orb, campos2021orb} and Structure-from-Motion (SFM) \cite{schonberger2016structure, cui2022vidsfm}. 
	As a broadly acknowledged matching pipeline, detector-based matching \cite{lowe2004distinctive, rublee2011orb, DeTone_2018_CVPR_Workshops, Dusmanu_2019_CVPR, revaud2019r2d2, tyszkiewicz2020disk, sarlin2020superglue, xia2022locality, chen2021learning, kuang2021densegap, shi2022clustergnn} is typically accomplished by (i) detecting and describing a set of sparse keypoints such as SIFT \cite{lowe2004distinctive}, ORB \cite{rublee2011orb}, and learning-based equivalents~\cite{detone2018superpoint, revaud2019r2d2}, (2) instituting point-to-point correspondences via nearest neighbour search or more advanced matching algorithms \cite{bian2017gms}. 
	The use of a feature detector narrows the matching search space, revealing the general efficacy of such detector-based matching process. 
	Nonetheless, when dealing with image pairs with severe viewpoint variations, such pipeline struggles to build reliable correspondences since the detectors are essentially incapable of extracting repeated keypoints in this case. 
	
	
	
	\begin{figure}
		\centering
		\includegraphics[width=0.99\hsize]{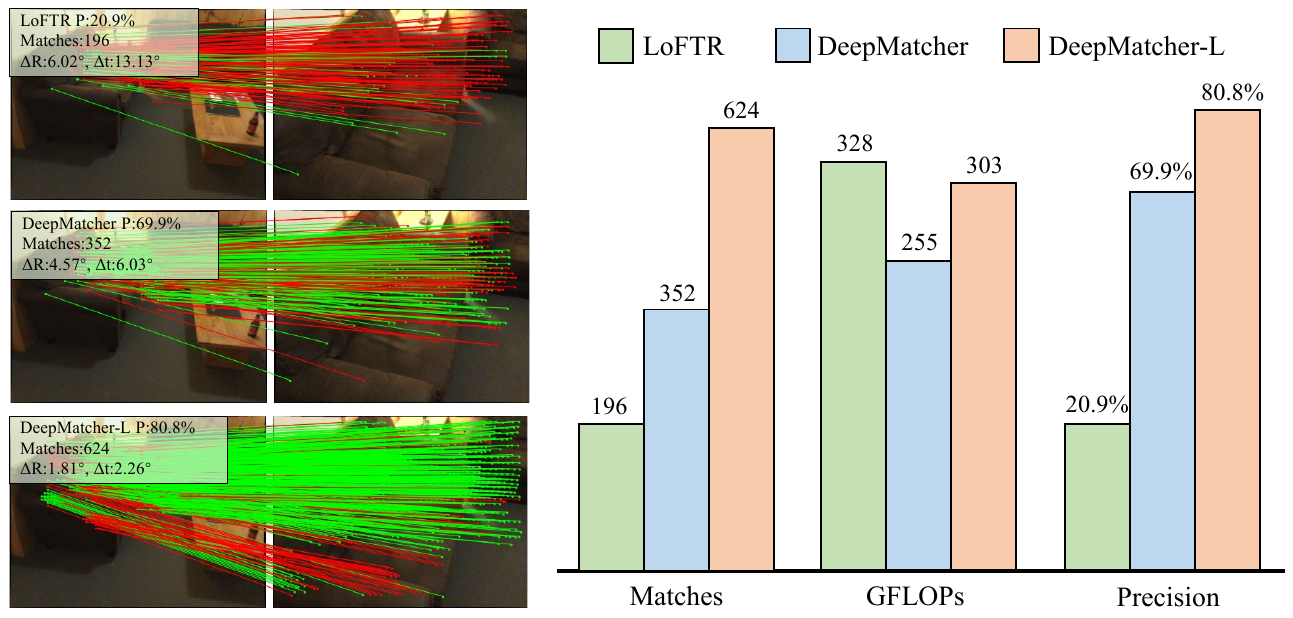}
		\caption{\textbf{The comparison between LoFTR and DeepMatcher under large viewpoint changes.} 
			DeepMatcher families considerably outperform LoFTR with more dense and precise matches while using less GFLOPs. 
		}
		\label{FIG1}
		\vspace{-2em}
	\end{figure}

	Parallel to the detector-based matching, another stream of research seeks to establish correspondences directly from original images by extracting visual descriptors on dense grids across an image, thus assuring that substantial repeating keypoints could well be captured \cite{rocco2018neighbourhood, rocco2020efficient, li2020dual, zhou2021patch2pix, efe2021dfm, sun2021loftr, wang2022matchformer, truong2022topicfm, chen2022aspanformer}. 
	Earlier detector-free matching works \cite{rocco2018neighbourhood, rocco2020efficient, li2020dual, zhou2021patch2pix, efe2021dfm} generally depend on iterative convolution based on correlation or cost volume to identify probable neighbourhood consensus. 
	Transformer~\cite{vaswani2017attention} has recently attracted considerable interest in computer vision due to its excellent model capability and superior potentials for capturing long-range relationships. 
	On the basis of this insight, various studies base their modelling of long-range relationships on Transformer backbone \cite{sun2021loftr, wang2022matchformer, truong2022topicfm, chen2022aspanformer}. 
	As a representative work, LoFTR \cite{sun2021loftr} updates features by repeatedly interleaving the self- and cross-attention layers, and replace vanilla Transformer with linear Transformer \cite{katharopoulos2020transformers} to achieve manageable computation cost. 
	These detector-free methdos can generate repeatable keypoints in indistinct regions with poor textures or motion blur, thus yielding impressive results. 
	After witnessing the success of detector-free methods, an intriguing issue arises: could we build a deeper yet compact local feature matcher to further improve performance while reducing computing costs? 
	Intuitively, when individuals match images, they scan the images back and forth, and the more times they scan them, the easier it is for them to remember the easier-to-match features, which indicates that a deep local feature matcher could display superior matching ability. 
	However, as demonstrated in LoFTR, doubling the number of Transformer layers has little effect on the results, which is contrary to expectations. 
	In this work, we argue that the following obstacles hinder us from developing a deep local feature matcher for detector-free methods:
	
	(i) Typical detector-free methods begin with a convolution neural network (CNN) as the basic feature extractor, followed by Transformer layers to capture long-range relevance so that generating credible correspondences. 
	In terms of context ranges, there is apparently a gap between the global receptive field of Transformer and the local neighborhood of CNN, which is detrimental to subsequent stages involving deep feature interaction. 
	
	
	(ii) The translation invariance of CNN causes ambiguity in scenes with recurring geometry patterns or symmetrical structures. 
	Current detector-free methods utilize absolute position encodings before Transformer layers to tackle this issue, while the position information would disappear as the Transformer layers grow deeper. 
	Moreover, humans naturally associate items across observations by referring to not only their absolute position but also their relative position. 
	
	(iii) Intuitively, the depth of the network is more prominent than width in the field of feature matching. 
	However, as the linear Transformer layer in LoFTR goes deeper, the model fails to learn effective context aggregation from deeper layers since the linear Transformer uses a context-agnostic manner to approximate self-attention, which cannot efficaciously simulate relevance among all keypoints. 
	
	
	To this end, we propose DeepMatcher, a deep local feature matching network that can produce more human-intuitive and simpler-to-match features for accurate correspondence with less computational complexity, as shown in \cref{FIG1}. 
	Firstly, we utilize a CNN network to generate pixel tokens with enriched features. 
	Secondly, an \textbf{Feature Transition Module (FTM)} is introduced to ensure a smooth transition from the locally aggregated features extracted by CNN to features with a global receptive field extracted by Transformer. 
	Then, we propose a \textbf{Slimming Transformer (SlimFormer)} to build deep network that strengthens long-range global context modelling intra-/inter-images. 
	Technically, SlimFormer leverages vector-based attention that efficiently handles pixel tokens with linear complexity for robust long-range global context aggregation.
	Besides, a relative position encoding is applied to each SlimFormer to clearly express relative distance information, boosting the network's capacity to convey information, particularly in deeper layers.
	Moreover, SlimFormer utilizes a layer-scale strategy that enables the network to assimilate message exchange from the residual block adaptively, thus allowing it to simulate the human behavior that human can receive different matching information each time they scan an image pair. 
	By interleaving the self- and cross-SlimFormer multiple times, DeepMatcher learns the discriminative features to construct dense matches at the coarse level by \textbf{Coarse Matches Module (CMM)}.
	Ultimately, we view the match refinement as a combination of classification and regression problems and devise \textbf{Fine Matches Module (FMM)} to predict confidence and offset concurrently, obtaining robust and accurate matches. 
	
	To summarize, the main contributions of this work are as follows:
	\begin{itemize}
		\item{We propose DeepMatcher, a deep Transformer-based network for local feature matching, achieving state-of-the-art results on various benchmarks.}
		\item{We propose a Feature Transition Module (FTM) to ensure a smooth transition from the locally aggregated features extracted by CNN to features with a global receptive field extracted by SlimFormer.}
		\item{We propose a Slimming Transformer (SlimFormer) that integrates long-range global context aggregation, relative position encoding, and layer-scale strategy to 
			enable DeepMatcher to be extended into dozen layers.}
		\item{We propose Fine Matches Module (FMM) that views the match refinement as a combination of classification and regression problems to optimze coarse matches, deriving robust and accurate matches.}
	\end{itemize} 
	
	\section{Related Work}
	\subsection{Detector-based Methods}
	The conventional pipeline of detector-based matching systems detects two sets of keypoints, describes them with high-dimensional vectors, and then implements a matching algorithm to generate matches between the two sets of keypoints~\cite{karpushin2016keypoint, balntas2016learning}.  
	
	Regarding feature detection and description, there are numerous handcrafted methods that seek to strike a balance between accuracy and efficiency~\cite{lowe2004distinctive, rublee2011orb, jiang2019robust}.
	However, the handcrafted descriptors are fragile when coping with image pairs with extreme appearance variations.
	With the development of deep learning \cite{sun2022proposal, fu2018refinet, song20166, fan2022seeing}, numerous approaches leverage elaborate convolution neural network (CNN) to extract robust feature representations, hence achieving superior performance. 
	SuperPoint \cite{DeTone_2018_CVPR_Workshops} builds a large dataset of pseudo-ground truth interest point locations in real images, supervised by the interest point detector itself, as opposed to a large-scale human annotation. 
	D2-Net \cite{Dusmanu_2019_CVPR} makes the collected keypoints more stable by delaying the detection to a later stage. 
	Subsequently, the aforementioned methods utilize the nearest neighbor search, followed by a robust estimator, such as RANSAC or its variants \cite{fischler1981random, fragoso2013evsac, barath2019magsac, mousavi2022two, fragoso2017ansac}, to find matches between the retrieved keypoints.

	Recent researches~\cite{sarlin2020superglue, chen2021learning, kuang2021densegap, shi2022clustergnn} has interpreted local feature matching as a graph matching problem involving two sets of features. 
	These methods utilize keypoints as nodes to construct graph neural network (GNN), employ the self- and cross-attention layers in Transformer to exchange global visual and geometric messages across nodes, and then generate the matches in accordance with soft assignment matrixes. 
	Typically, SuperGlue \cite{sarlin2020superglue} utilizes self- and cross-attention in Transformer to integrate global context information, followed by the Sinkhorn algorithm to generate matches according to the soft assignment matrix. 
	Nonetheless, the matrix multiplication in vanilla Transformer results in quadratic complexity with respect to the number of keypoints, making SuperGlue costly to deal with substantial keypoints. 
	To tackle this problem, many approaches attempt to ameliorate the structure of SuperGlue. 
	SGMNet \cite{chen2021learning} exploits the sparsity of graph neural network to lower the computation complexity. 
	ClusterGNN \cite{shi2022clustergnn} employs a progressive clustering module adaptively to divide keypoints into different subgraphs to reduce computation. 
	However, limited to inherent essence, detector-based approaches are incapable of extracting repeated keypoints when handling image pairs with large appearance variations.

	\subsection{Detector-free Methods}
	Detector-free methods exclude the feature detector and generate dense matches directly from the original images. 
	Earlier detector-free matching researches \cite{rocco2018neighbourhood, rocco2020efficient, li2020dual, zhou2021patch2pix, efe2021dfm} generally utilize convolutional neural network (CNN) based on  correlation or cost volume to identify probable neighbourhood consensus. 
	DRC-Net \cite{li2020dual} generates a 4D correlation tensor from the coarse-resolution features, which is refined by a learnable neighborhood consensus module to generate matches. 
	Patch2Pix \cite{zhou2021patch2pix} proposes a weakly supervised approach to learn matches that are consistent with the epipolar geometry of image pairs. 
	DFM \cite{efe2021dfm} uses pre-trained VGG architecture as a feature extractor and captures matches without any additional training strategy. 
	Although elevating the matching accuracy, these methods extract ambiguous feature representations and fail to discriminate incorrect matches owing to the limited receptive field of CNN. 
	
	To handle this issue, LoFTR \cite{sun2021loftr}, the pioneering detector-free GNN method, utlizes Transformer to realize global context information exchange and extracts matches in a coarse-to-fine manner. 
	Matchformer \cite{wang2022matchformer} proposes a human-intuitive extract-and-match scheme that interleaves self- and cross-attention in each stage of the hierarchical encoder. 
	Such a match-aware encoder releases the overloaded decoder and makes the model highly efficient. 
	QuadTree \cite{tang2022quadtree} proposes a novel Transformer structure that builds token pyramids and computes attention in a coarse-to-fine manner.
	Then, the QuadTree Transformer is integrated into LoFTR and achieves superior matching performance. 
	TopicFM \cite{truong2022topicfm} applies a topic-modeling strategy to encode high-level contexts in images, which improves the robustness of matching by focusing on the same semantic areas between the images.
	ASpanFormer \cite{chen2022aspanformer} proposes a Transformer-based detector-free architecture, in which the flow maps are regressed in each cross-attention phase to perform local attention. 
	These detector-free methods are capable of generating repeatable keypoints in indistinct regions with poor textures or motion blur, thus resulting in amazing results. 
	However, the architecture of existing detector-free methods is designed as shallow-broad, and building a deeper and more compact local feature matcher to further improve performance while reducing computational costs has not been investigated. 

	\subsection{Efficient Transformer}
	In the vanilla Transformer, the memory cost is quadratic to the length of sequences due to the matrix multiplication, which has become a bottleneck for Transformer when dealing with long sequences. 
	Recently, several approaches have been proposed to improve the efficiency of Transformer \cite{katharopoulos2020transformers, zaheer2020big, wu2021fastformer}. 
	Linear Transformer \cite{katharopoulos2020transformers} expresses self-attention as a linear dot product of kernel feature maps and makes use of the associativity property of matrix products to reduce the computational complexity. 
	BigBird \cite{zaheer2020big} combines local attention and global attention at certain positions and utilizes random attention on several randomly selected token pairs. 
	FastFormer \cite{wu2021fastformer} uses additive attention mechanism to model global contexts, achieving effective context modeling with linear complexity. 
	

	\begin{figure}
		\centering
		\includegraphics[width=0.99\hsize]{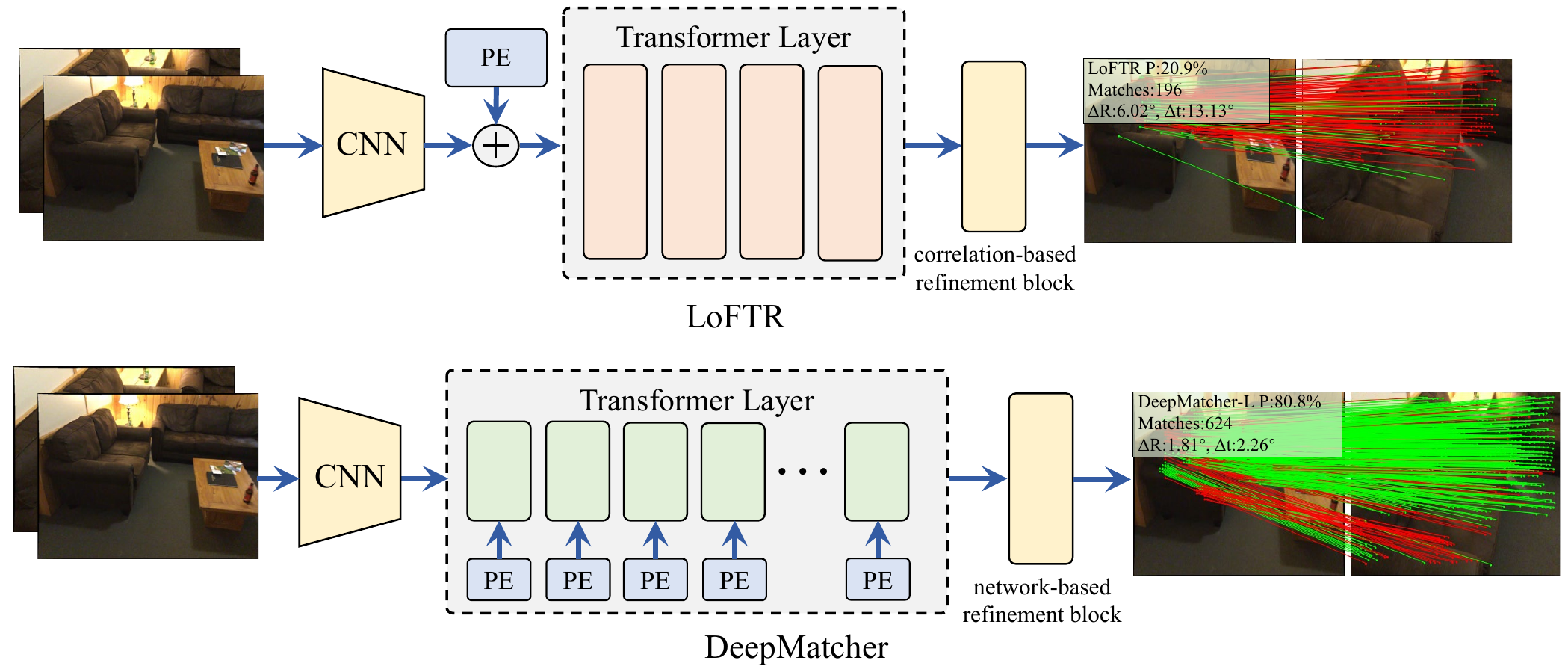}
		\caption{\textbf{The comparison between LoFTR and DeepMatcher.} 
			DeepMatcher designs a deep-narrow Transformer layers to capture more human-intuitive and simpler-to-match features.
			Besides, the position encoding (PE) is integrated to each Transformer layer to convey position information in deep layers.
			Moreover, a network-based refinement block is proposed to extract more precise matches.
		}
		\label{att_layer}
		\vspace{-10pt}
	\end{figure}
	\begin{figure*}
		\centering
		\includegraphics[width=1.00\hsize]{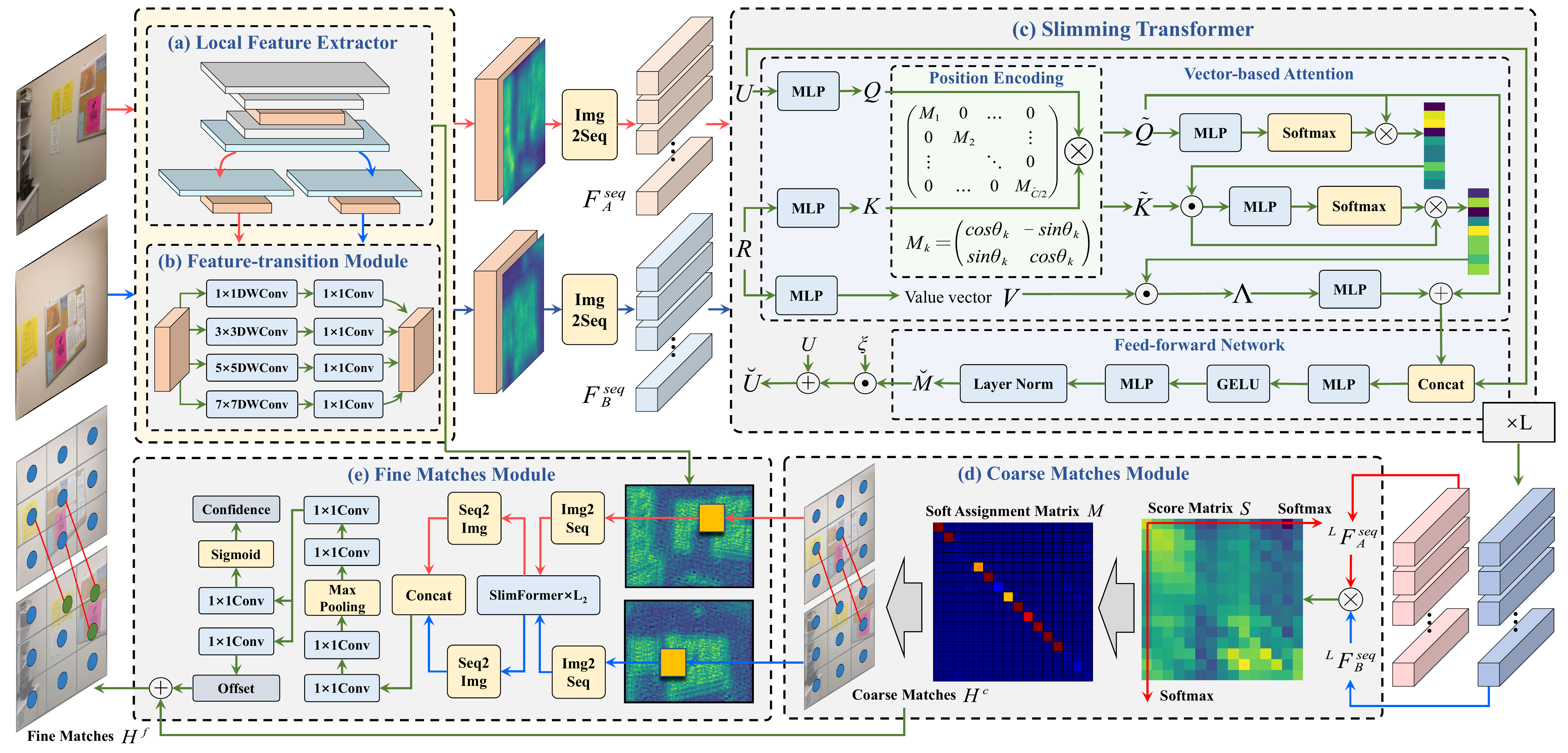}
		\caption{\textbf{The network architecture of DeepMatcher.} DeepMatcher takes an image pair$(I_{A}, I_{B})$ as input and generates transitional features from \textit{Local Feature Extractor} and \textit{Feature Transition Module}. Then, DeepMatcher interleaves \textit{Slimming Transformer} by $L$ times to perform long-range context aggregation.
			\textit{Coarse Matches Module} is utilized to establish coarse matches, which are optimize to fine matches by \textit{Fine Matches Module}.
		}
		\label{overall}
		\vspace{-1em}
	\end{figure*}
	
	\section{Methodology}
	\subsection{Overall}
	Intuitively, when humans match images, they scan the images back and forth, and the more times they scan them, the simpler it is for them to recall the easier-to-match features, which suggests that a deep local feature matcher can exhibit higher matching abilities. 
	Thus, as shown in \cref{att_layer}, we consider depth of the network is more prominent than width and present a deep Transformer-based network, namely DeepMatcher. 
	As shown in \cref{overall}, given the image pair $I_A$ and $I_B$, our network produces reliable and accurate matches across images in an end-to-end manner. 
	The matching process starts with a CNN-based encoder to extract the fine-level features $\Bar{F}_{A}, \Bar{F}_{B}$ and coarse-level features $\hat{F}_{A}, \hat{F}_{B}$. 
	Before feeding these features to Slimming Transformer (SlimFormer), we utilize the Feature Transition Module (FTM) to guarantee a smooth transition to SlimFormer. 
	Then, we utilize SlimFormer to achieve long-range global context aggregation intra-/inter-images in an efficient and effective way.  
	A relative position encoding is applied on each SlimFormer to explicitly model relative distance information, hence enhancing the DeepMatcher's ability to convey information, particularly in deeper layers. 
	A layer-scale strategy is also leveraged in each SlimFormer to enables the network to assimilate message exchange from the feed-forward modules adaptively, thus allowing it to simulate the human behavior that human can receive different matching information each time they scan an image pair. 
	After interleaving SlimFormer by $L$ times, the enhanced features $^{L}F^{seq}_A$ and $^{L}F^{seq}_B$ are utilized to establish coarse matches $H_{c}$, which are further optimized to fine matches $H_{f}$ using Correspondence Refine Module (CRM).
	In the following part, we introduce the details and underlying insights of each individual block.
	
	\subsection{Local Feature Extractor}
	As the first part of DeepMatcher, we use a standard convolutional neural network (CNN) with FPN \cite{lin2017feature} to extract the coarse-level features $\hat{F_A}$, $\hat{F_B} \in \mathbb{R}^{\hat{C} \times H/8 \times W/8}$, and the fine-level features $\bar{F}_A$, $\bar{F}_B  \in \mathbb{R}^{\bar{C} \times H/2 \times W/2}$ for the image pair $I_A$ and $I_B$, where $H$ and $W$ are the height and width of the original images, $\hat{C}$, $\bar{C}$ denote feature dimension.
	For convenience, we denote $N = H/8 \times W/8$ as the number of pixel tokens. Since each pixel in $\hat{F}_A$, $\hat{F}_B$ represents an $8 \times 8$ grid in the original images $I_{A}, I_{B}$, we view the central position of all grids as the pixel coordinates $P_{A}, P_{B} \in \mathbb{R}^{N \times 2}$ of keypoints.
	
	
	\subsection{Feature Transition Module (FTM)}
	In the subsequent steps, we construct graph neural network (GNN) and propose SlimFormer that leverages self-/cross-attention in Transformer to aggregate global context information intra-/inter-image. 
	Nevertheless, there is apparently a gap between the feature extractor and SlimFormer in terms of context ranges, which is deleterious to subsequent steps involving deep feature interaction. 
	Besides, representing features at multiple scales is so critical for discriminating objects or regions of varying sizes that it can ensure prominent features at various scales can be preserved for deep features aggregation. 
	Thus, we propose a Feature Transition Module (FTM) inserted between the local feature extractor and SlimFormer to adjust the receptive fields of the extracted features, ensuring effective deep feature interaction in SlimFormer. 
	Specifically, instead of directly using $(1\times1, 3\times3, 5\times5, 7\times7)$ convolution, FTM adopts $(1\times1, 3\times3, 5\times5, 7\times7)$ depth-wise convolution \cite{howard2017mobilenets} followed by $1\times1$ point-wise convolution to reduce model parameters and computation, obtaining multi-scale feature representations.
	Then, we concatenate the features along channel dimension, hence enlarging the receptive fields of $\hat{F_A}, \hat{F_B}$.
	Finally, we derive the updated  $F_{A}^{ftm}, F_{B}^{ftm} \in \mathbb{R}^{\hat{C} \times H/8 \times W/8}$, which can be formulated as:
	\begin{equation}
		\begin{aligned}
			FTM&(F) = [C_{1}^{1/4}(DW_{1}(F))||C_{1}^{1/4}(DW_{3}(F))||\\
			& C_{1}^{1/4}(DW_{5}(F))||C_{1}^{1/4}(DW_{7}(F))], \\
			F_{A}^{ftm} & = FTM(\hat{F}_{A}),  \ \ \ F_{B}^{ftm} = FTM(\hat{F}_{B}),
		\end{aligned}
	\end{equation}
	where $C_{1}^{1/4}$ means using $1 \times 1$ convolution to squeeze the channel dimension to $\hat{C} / 4$;
	$DW_{1}, DW_{3}, DW_{5}, DW_{7}$ mean depth-wise convolution with kernel size of 1, 3, 5, 7, respectively;
	$[\cdot||\cdot]$ means concatenation along the channel dimension.

	\subsection{Slimming Transformer (SlimFormer)}
	We flatten the updated enhanced features $F_{A}^{ftm}, F_{B}^{ftm}$ to be the input sequence for deep feature aggregation, obtaining $F_{A}^{seq}, F_{B}^{seq} \in \mathbb{R}^{N \times \hat{C}}$. 
	Following \cite{sarlin2020superglue}, we view keypoints with features $F_{A}^{seq}, F_{B}^{seq}$ in image pairs as nodes to construct GNN, in which the global context aggregation intra-/inter-image is performed.
	Intuitively, more observations between images can result in more precise matches, indicating that deep feature interaction is essential for local features matching task. 
	Nevertheless, the ablation study of LoFTR demonstrates that the matching performance has not been significantly improved with more Transformer layers. 
	We attribute this phenomenon to the following reasons:
	(i) LoFTR only utilizes absolute position encoding before Transformer layers, where the position information would disappear when the Transformer layers grow deeper. 
	Moreover, humans primarily associate objects by referring to their relative positions. 
	(ii) The linear Transformer utilized in LoFTR uses a context-agnostic manner to approximate self-attention, which cannot fully model relevance among all keypoints, especially in deep layers. 
	To handle this dilemma, we propose SlimFormer that leverages relative position information and global context information to boost the capability of DeepMatcher to convey abundant information, hence extracting discriminative feature representations $^{L}F^{seq}_A, ^{L}F^{seq}_B \in \mathbb{R}^{\hat{C} \times H/8 \times W/8}$.

	
	\textbf{Vector-based Attention (VAtt) Layer.}
	Instead of using a context-agnostic manner to approximate self-attention, we convert query vector to global query contexts and leverage element-wise product to model relevance among all keypoints.
	Technically, during each feature enhancement process, we utilize self-/cross-attention to aggregate long-range context information intra-/inter-images.
	For self-attention, the input features $U$ and $R$ are same (either $(F_{A}^{seq}, F_{A}^{seq})$ or $(F_{B}^{seq}, F_{B}^{seq})$).
	For cross-attention, the input features $U$ and $R$ are different (either $(F_{A}^{seq}, F_{B}^{seq})$ or $(F_{B}^{seq}, F_{A}^{seq})$).
	Firstly, SlimFormer transforms the input features $U$ and $R$ into the query, key, and value vectors $Q, K, V \in \mathbb{R}^{N \times \hat{C}}$.
	\begin{equation}
		\begin{split}
			Q = U W_Q, \ \ \ K = R W_K, \ \ \ V = R W_V,
		\end{split}
	\end{equation}
	where $W_Q$, $W_K$, $W_V \in \mathbb{R}^{\hat{C} \times \hat{C}}$ denote learnable weights for feature transformation.
	Then, we perform relative position encoding on query vector $Q$ and key $K$.
	\begin{equation}
		\Tilde{Q} = DPE(Q), \ \ \ \Tilde{K} = DPE(K),
		\label{eq3}
	\end{equation}
	where $DPE(\cdot)$ means relative position encoding operation, described below.
	
	Next, modeling the context information of the input features based on the interactions among $\Tilde{Q}, \Tilde{K}$, and $V$ is a critical problem for Transformer-like architectures. 
	In the vanilla Transformer, dot-product attention mechanism leads to quadratic complexity, making it unrealistic to establish deep Transformer layers.
	A potential method to reduce the computational complexity is to summarize the attention matrices before modeling their interactions. 
	Inspired by \cite{wu2021fastformer}, we introduce vector-based attention that effectively models long-range interactions among pixel tokens to alleviate this bottleneck. 
	Instead of computing a quadratic attention map $QK^{T}$ that encodes all possible interactions between candidate matches, we form a compact representation of query-key interactions via vector-based attention that computes the correlation between global query vector and each key vector.
	Specifically, we firstly leverage MLP to calculate the weight $\Tilde{Q}_{imp} \in \mathbb{R}^{1 \times N}$ of each query vector:
	\begin{equation}
		\Tilde{Q}_{imp} = Softmax(MLP(\Tilde{Q})),
		\label{Qweight}
	\end{equation}
	where $Softmax(\cdot)$ means softmax operation.
	
	The global query vector $\Breve{Q} \in \mathbb{R}^{1 \times \hat{C}}$ is set to be a linear combination of $\Tilde{Q}$:
	\begin{equation}
		\Breve{Q} = \Tilde{Q}_{imp} \otimes \Tilde{Q}
	\end{equation}
	where $\otimes$ means matrix multiplication.
	
	Then, we utilize the element-wise multiplication between the global query vector $\Breve{Q}$ and each key vector to model their interaction, obtaining context-aware key vector $\Tilde{K}_{Q} \in \mathbb{R}^{N \times \hat{C}}$:
	\begin{equation}
		\Tilde{K}_{Q} = \Breve{Q} \odot \Tilde{K},
	\end{equation}
	where $\odot$ denotes element-wise multiplication.
	
	We utilize a similar vector-based attention to extract global context-aware key vector $\Breve{K}_{Q}$ and model the interaction between $\Breve{K}_{Q}$ and $V$:
	\begin{equation}
		\begin{aligned}
			\Tilde K_{Qimp} = S&oftmax(MLP(\Tilde{K}_{Q})) \\
			\Breve{K}_{Q} =& \Tilde K_{Qimp} \otimes \Tilde{K}_{Q} \\
			\Lambda &= \Breve{K}_{Q} \odot V
		\end{aligned}
		\label{KQweight}
	\end{equation}
	
	Subsequently, we employ a MLP and short-cut structure to derive the global message $M \in \mathbb{R}^{N \times \hat{C}}$.
	\begin{equation}
		M = MLP(\Lambda) + \Tilde{Q}
	\end{equation}
	
	For convenience, we define the process of vector-based attention layer as:
	\begin{equation}
		M = \mathit{VAtt}(U, R)
	\end{equation}
	
	\textbf{Feed-forward Network (FFN).}
	Inspired by conventional Transformers, we employ a feed-forward network applied to $M$ to extract discriminative features for effectively deep features aggregation.
	The feed-forward network consists of two fully-connected layers and a GELU activation function.
	The hidden dimension between the two fully-connected layers is extended by a scale rate $\gamma$ to learn abundant feature representation.
	This process can be formulated as:
	\begin{equation}
		\begin{aligned}
			FFN(U, M) = MLP_{1 / \gamma}(GELU(
			MLP_{\gamma / 2}([U||M]))),
		\end{aligned}
		\label{ffn}
	\end{equation}
	where $MLP_{1 / \gamma}, MLP_{\gamma / 2}$ mean expand the channel dimension by $1 / \gamma, \gamma / 2$ times with a MLP, respectively;
	$[\cdot||\cdot]$ means concatenation along channel dimension;
	$GELU(\cdot)$ means GELU activation function.
	Ultimately, we obtain enhanced message $\Breve{M} \in \mathbb{R}^{N \times \hat{C}}$.
	
	\textbf{Layer Scale Strategy.}
	Intuitively, people obtain different message after observing images each time, which inspires us to propose a layer-scale strategy.
	Specifically, in accordance with ResNet \cite{he2016deep}, we utilize a shortcut structure to realize efficient training.
	Then, we design a learnable scaling factor $\xi$ to adaptively balance original features $U$ and enhanced message $\Breve{M}$, which is formulated as.
	\begin{equation}
		\Breve{U} = U + \xi \Breve{M}
		\label{learnable_scale}
	\end{equation}
	By incorporating $\xi$ into SlimFormer, SlimFormer can easily simulate the human behaviour that humans acquire different matching cues each time they scan an image pair. 
	
	\textbf{Relative Position Encoding (RPE).}
	The local feature extractor learns strict translation invariant features, which could cause ambiguity in scenes that have repetitive geometry texture or symmetric structures.
	Previous works \cite{sarlin2020superglue, sun2021loftr, chen2021learning, wang2022matchformer} attach a distinctive absolute positional embedding to each keypoint, thus alleviating such ambiguity.
	However, compared with absolute position, relative position is more conducive for humans to establish connections between objects.
	Therefore, we argue that incorporating the explicit relative position dependency during each deep feature aggregation is essential for distinguishing identical features. 
	However, relative position is not applicable to transformers with linear complexity as they do not explicitly calculate the quadratic complexity attention matrix. 
	To this end, we employ rotary positional embedding (RoPE) \cite{su2021roformer} that leverages absolute position encoding to achieve relative position encoding without manipulating the attention matrix. 
	Given a pixel token $T_i$ and its features $F_i \in \mathbb{R}^{\hat{C}}$, the rotary position encoding function is defined by:
	
	\begin{equation}
		\begin{split}
			Pos(T_i, F_i) = \Theta(T_i)F_i = \begin{pmatrix}
				M_1 &0   &\dots  &0 \\
				0&  M_2  &  &\vdots \\
				\vdots&  &\ddots    &0 \\
				0&\dots  &0    & M_{\hat{C}/2}
			\end{pmatrix} F_i,
		\end{split}
	\end{equation}
	where $\Theta(T_i) \in \mathbb{R}^{\hat{C} \times \hat{C}}$ is a block diagonal matrix. 
	Each block with size of $2\times2$ is defined by:
	\begin{equation}
		\begin{split}
			\begin{aligned}
				M_k =  \begin{pmatrix}
					cos \ i\theta _{k} & -sin \ i\theta _{k}  \\
					sin \ i\theta _{k} & cos \ i\theta _{k} \\
				\end{pmatrix}, \ \ \ 
				\theta_k = \frac{1}{10000^{2(k-1)/\hat{C}}}
			\end{aligned}
		\end{split}
	\end{equation}
	where $\theta_k$ encodes the index of the feature channel.
	
	Compared to sinusoidal encoding \cite{sun2021loftr, wang2022matchformer, sarlin2020superglue}, rotary positional embedding has two advantages: (i) $\Theta(\cdot)$ is an orthogonal function, the encoding only changes the feature’s direction but not the feature's length, which could stabilize the learning process. (ii) The dot product of two encoded features $<Pos(T_i, F_i), Pos(T_j, F_j)>$ in self-attention of vanilla Transformer can be derived to: 
	\begin{equation}
		\begin{split}
			[\Theta(T_i)F_i]^T \Theta(T_j)F_j = (F_i)^T \Theta(T_j - T_i) F_j
		\end{split}
	\end{equation}
	which means the relative 2D distance information can be explicitly revealed by the dot product. 
	
	Since RoPE injects position information by rotation, which maintains the norm of hidden representations unchanged, such positional encoding can be directly applied to linear complexity transformers as demonstrated in \cite{su2021roformer}. 
	In SlimFormer, we implement this by employing rotary positional embedding into $Q, K$ to incorporate relative position information, as illustrated in \cref{overall} or \cref{eq3}.
	For more details about RoPE, we encourage readers to refer to original papers.
	

	\textbf{Self-/Cross-SlimFormer.}
	In summary, the SlimFormer is formatted as:
	\begin{equation}
		Slim(U, R) = U + \xi FFN(U, \mathit{VAtt}(U,R))
	\end{equation}
	
	We perform $L$ times of SlimFormer for feature enhancement.
	During the $l$-th feature enhancement, we use self-/cross-attention mechanism to integrate intra-/inter-image information, which can be formulated as:
	\begin{equation}
		\begin{aligned}
			^{l-1}F^{seq}_{A} &= Slim(^{l-1}F^{seq}_{A},\ ^{l-1}F^{seq}_{A}), \\
			^{l-1}F^{seq}_{B} &= Slim(^{l-1}F^{seq}_{B},\ ^{l-1}F^{seq}_{B}), \\
			^{l}F^{seq}_{A} &= Slim(^{l-1}F^{seq}_{A},\ ^{l-1}F^{seq}_{B}), \\
			^{l}F^{seq}_{B} &= Slim(^{l-1}F^{seq}_{B},\ ^{l}F^{seq}_{A})
		\end{aligned}
	\end{equation}
	
	Ultimately, we incorporate relative position information and global context message into enhanced features $^{L}F^{seq}_A, ^{L}F^{seq}_B$.
	
	\subsection{Coarse Matches Module (CMM)}
	Given $^{L}F^{seq}_A$ and $^{L}F^{seq}_B$, we utilize inner product of $^{L}F^{seq}_A,\ ^{L}F^{seq}_B$ to calculate the score matrix $S\in\mathbb{R}^{N \times N}$.
	\begin{equation}
		\begin{aligned}
			S(i,j) &= \langle ^{L}F^{seq}_A, \ ^{L}F^{seq}_B  \rangle,
		\end{aligned}
	\end{equation}
	where $\langle \cdot , \cdot \rangle$ means the inner product.
	Subsequently, we apply softmax operator on both dimensions (denoted as dual-softmax operation) to convert the $S$ to soft assignment matrix $G \in \mathbb{R}^{N \times N}$:
	\begin{equation}
		\begin{aligned}
			G = Softmax(S)_{col} \cdot Softmax(S)_{row},
		\end{aligned}
	\end{equation}
	where $Softmax(\cdot)_{col}, \ Softmax(\cdot)_{row}$ mean performing softmax on each column and row of $S$, respectively.
	
	Then, for the $i$-th keypoint in $I_{A}$ and the $j$-th keypoint in $I_{B}$, we regard them as a pair of predicted coarse matches if they satisfy the following two conditions:
	(i) The soft assignment score is higher than a predefined threshold $\lambda$: $G(i,j)>\lambda$.
	(ii) They satisfy the mutual nearest neighbor (MNN) criteria, i.e., $G(i, j)$ is the maximum value in the corresponding row and column.
	Ultimately, we derive the index $D$ of anchor points in coarse matches:
	\begin{equation}
		\begin{aligned}
			D = \{(i,j)|(i,j) \in {\rm MNN}(M), M(i,j)>\lambda \}
		\end{aligned}
		\label{coarse_equa}
	\end{equation}
	
	Given the index $D$ and the keypoints coordinates $P_{A}, P_{B}$, the coarse matches $H^{c} = \{(P_{A}^{c}, P_{B}^{c})\}$ are formulated as:
	\begin{equation}
		\begin{aligned}
			H^{c} = \{(P_{A}(i),P_{B}(j))\ | \ \forall(i,j) \in D \}
		\end{aligned}
	\end{equation}

	\begin{figure}
		\centering
		\includegraphics[width=0.99\hsize]{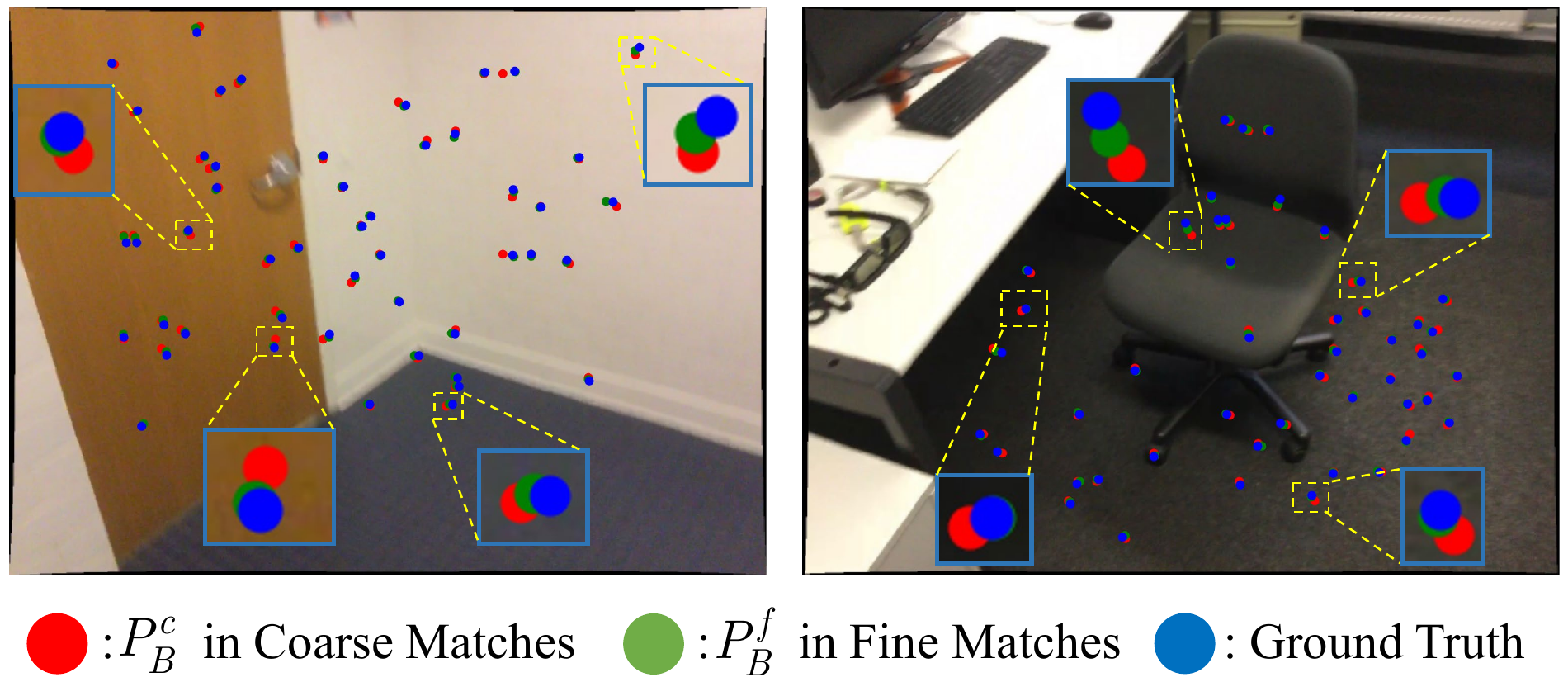}
		\caption{\textbf{Visualization of refinement result.} The keypoints (green) in the fine matches approximate the ground-truth (blue).} 
		\label{refinement}
		\vspace{-15pt}
	\end{figure}
	\subsection{Fine Matches Module (FMM)}

	After establishing coarse matches, a coarse-to-fine module is applied to refine these matches to the original picture resolution.
	However, the coarse-to-fine module in LoFTR only predicts the offset of the coarse matches without appraising whether the predicted matches are reliable.
	To tackle this issue, we view the match refinement as a combination of classification and regression problems and design Fine Matches Module to predict confidence and offset concurrently.
	
	As shown in \cref{overall}, for each coarse match, we locate its position at fine-level feature maps and crop two sets of local image patches with the size of $w \times w$, obtaining local features $\bar{F}_{A}^{w}, \bar{F}_{B}^{w} \in \mathbb{R}^{K \times \bar{C} \times w \times w}$, where $K$ is the number of coarse matches.
	Then, we flatten $\bar{F}_{A}^{w}, \bar{F}_{B}^{w}$ to be sequences, implement SlimFormer to perform $L_2$ times of global information passing, and rearrange the sequences into 2D feature maps, obtaining $^{L_{2}}\bar{F}_{A}^{w}, ^{L_{2}}\bar{F}_{B}^{w}$.
	The feature maps are concatenated along channel dimension and fed into a network, which is comprised of two convolution layers, a max pooling layer, and four convolution layers.
	The network predicts the offset $\Delta \in \mathbb{R}^{K \times 2}$ of the $P_B^{c}$ and the confidence $c \in \mathbb{R}^{K \times 1}$ of the predicted coarse matches:
	\begin{equation}
		\begin{aligned}
			\bar{F}_{mid} &=  C_{1}(C_{1}(P_{max}(C_{1}(C_{1}([^{L_{2}}\bar{F}_{A}^{w}||^{L_{2}}\bar{F}_{B}^{w}]))))), \\[1pt]
			&c = Sig(C_{1}(\bar{F}_{mid})), \ \ \ \Delta = C_{1}(\bar{F}_{mid}),
		\end{aligned}
	\end{equation}
	where $P_{max}$ means global max pooling operation; $C_{1}(\cdot)$ means $1\times1$ convolution; $[\cdot||\cdot]$ denotes concatenation along the channel dimension; $Sig(\cdot)$ means sigmoid function.
	
	Ultimately, we obtain the fine matches $H^{f} = \{(P_{A}^{f}, P_{B}^{f})\}$:
	\begin{equation}
		\begin{aligned}
			H^{f} = \{(P_{A}^{c}(i),  P_{B}^{c}(i)+\Delta(i)) \ | \ i \in \{1,2,3,...,K \}\}
		\end{aligned}
	\end{equation}
	
	\subsection{Loss}
	DeepMatcher generates final dense matches according to soft assignment matrix $G$ and offset $\Delta$.
	Therefore, the total loss $L^{all}$ of DeepMatcher comprises of matching loss $L^{m}$, regression loss $L^{r}$, and classification loss $L^{c}$.
	\begin{equation}
		L^{all} = L^{m} + \beta L^{r} + \phi L^{c},
	\end{equation}
	where $\beta$ and $\phi$ are weighting coefficient.
	
	\textbf{Matching Loss.}
	Following \cite{sun2021loftr}, we calculate the index $E^{gt}$ of the ground truth matches, which are utilized in conjunction with soft assignment matrix $G$ to calculate matching loss $L^{m}$ defined as focal loss \cite{lin2017focal}.
	\begin{equation}
		\begin{split}
			L^{m}&= -[\frac{1}{|E^{gt}|}\sum_{(i,j)\in E^{gt}}\alpha(1-G(i,j))^{\eta}log \ G(i,j)+ \\
			& \frac{1}{N-|E^{gt}|}\sum_{(i,j)\notin E^{gt}}(1-\alpha)G(i,j)^{\eta}log \ (1-G(i,j))],
		\end{split}
	\end{equation}
	where $\alpha$ is a weighting factor; $\eta$ is a focusing parameter; $|E^{gt}|$ means the number of ground truth matches.
	
	\textbf{Regression Loss.}
	For predicted matches $\{(P_{A}^{f}, P_{B}^{f})\}$, we project $P_{A}^{f}$ in the first image to second image, deriving $P_{B}^{gt}$.
	Then, the ground truth offset $\Delta^{gt}$ is formulated as:
	\begin{equation}
		\Delta^{gt} = P_{B}^{gt} - P_{B}^{f}
	\end{equation}
	
	According to predicted offset $\Delta$ and ground truth offset $\Delta^{gt}$, we define the regression loss $L^{r}$ as:
	\begin{equation}
		\begin{split}
			L^{r} = \frac{1}{K}\sum_{i=1}^{K}\Vert \Delta^{gt}(i)-\Delta(i) \Vert_2^2,
		\end{split}
	\end{equation}
	where $K$ is the number of predicted matches. 
	Notably, we ignore the predicted matches with $\Delta^{gt}$ larger than predefined threshold $\psi$.
	
	\textbf{Classification Loss.}
	For the predicted matches with ground truth offset less than $\psi$, we regard them as positive and define the classification label as $1$, while other matches are viewed as negative.
	Ultimately, we obtain the ground truth confidence $c^{gt}$, while are utilized to calculate classification loss $L^{c}$ together with predicted confidence $c$.
	\begin{equation}
		L^{c} = -\frac{1}{K}\sum_{i=1}^{K}\left[c^{gt}(i)log \ c(i) + (1-c^{gt}(i))log \ (1-c(i))\right]
	\end{equation}

	\section{EXPERIMENTS}
	
	\subsection{Implementation Details}
	\textbf{Architecture details.} 
	We adopt a slightly modified ResNet-18 with FPN for local feature extraction. 
	We use a width of 96 for the stem layer, followed by widths of [96, 128, 192] for the next three stages.
	We construct the FPN with levels $P_1$ through $P_3$ and take $P_3$ features as the coarse-level features, $P_1$ features as the fine-level features. 
	Thus, the dimensions of fine-level and coarse-level feature maps are $\overline{C} = 96, \ \hat{C} = 192$, respectively.
	The scale rate $\gamma$ in feed-forward network is set to $4$. 
	Following SuperGlue, we set the confidence threshold $\lambda=0.2$ to obtain coarse matches.
	Besides, we choose $w=5$ to crop local windows in fine-level feature maps for matches refinement.
	To reconcile the coarse matching loss, regression loss, and classification loss, we set both weighting coefficients $\beta$ and $\phi$ to $0.2$.
	For matching loss, we set the weighting factor $\alpha=0.25$ and the focusing parameter $\eta=2$.
	When making classification labels, we set $\psi$ to $8$. 
	In this work, we elaborately design two versions of DeepMatcher that interleave SlimFormer by $L = 6, 10$ times for feature enhancement, resulting in \textbf{DeepMatcher} and \textbf{DeepMatcher-L}. 
	
	\textbf{Training scheme for Scannet~\cite{dai2017scannet}.}
	We train DeepMatcher on Scannet~\cite{dai2017scannet} dataset with 32 Tesla V100 GPUs for indoor local feature matching. 
	In accordance with LoFTR, we sample 200 image pairs per scene at each epoch and balance scene variants over iterations. 
	We employ the AdamW solver for optimization with a weight decay of 0.1. 
	The initial learning rate is set to 6 $\times 10^{-4}$ and will decrease by 0.5 every 3 epochs. 
	We use gradient clipping that is set to 0.5 to avoid exploding gradients. 
	
	\textbf{Training scheme for MegaDepth~\cite{li2018megadepth}.}
	We train DeepMatcher on MegaDepth~\cite{li2018megadepth} datasets with 32 Tesla V100 GPUs for outdoor local feature matching. 
	Following LoFTR, we randomly sample 100 pairs from each sub-scene during each epoch of training. 
	We train DeepMatcher for 30 epochs in total. 
	We also employ the AdamW solver for optimization with a weight decay of 0.1. 
	The initial learning rate is set to 8 × $10^{-4}$, with a linear learning rate warm-up in 3 epochs from 0.1 to the initial learning rate. 
	We decay the learning rate by 0.5 every 4 epochs
	starting from the 4-th epoch. 
	
	\subsection{Indoor Pose Estimation}
	Typically, indoor pose estimation task is hampered by motion blur and significant viewpoint shifts. 
	There are commonly extensive regions of low textures in indoor scenes. 
	To evaluate the performance of DeepMatcher in such situations, we conducted indoor pose estimation experiments.
	
	\textbf{Dataset.} We use ScanNet~\cite{dai2017scannet} dataset to validate the effectiveness of DeepMatcher on indoor pose estimation task. 
	ScanNet consists of 1513 RGB-D sequences with RGB images and corresponding ground-truth poses in indoor environments. 
	Following \cite{sarlin2020superglue}, we select 230M image pairs with overlap values ranging from $0.4$ to $0.8$ as the training set and $1500$ image pairs as the test set. 
	All images are resized to $640 \times 480$.
	
	\textbf{Evaluation Protocol.}
	In accordance with \cite{sun2021loftr, sarlin2020superglue}, we report the area under the cumulative curve (AUC) of pose errors at the thresholds $(5^{\circ}, 10^{\circ}, 20^{\circ})$, where pose errors are defined as the maximum of translational and rotational errors between ground-truth poses and predicted poses by DeepMatcher. 
	Specifically, given the predicted dense matches, we utilize OPENCV to calculate the essential matrix $E$ and relative pose $\tilde{T}$ of image pairs.
	Then, the pose errors $\Delta T$ are defined as the maximum of translational and rotational errors between ground-truth relative pose $T=[R|t]$ and estimated relative pose $\tilde{T}=[\tilde{R}|\tilde{t}]$:
	\begin{equation}
		\begin{aligned}
			\Delta T =& max(\Delta t, \Delta R), \\
			\Delta t = arccos(\frac{\tilde{t} \cdot t}{||\tilde{t}||_{2} \cdot ||t||_{2}}),& \ \
			\Delta R = arccos(\frac{tr(\tilde{R}^{T}R)-1}{2}),
		\end{aligned}
	\end{equation}
	where $\Delta t$ and $\Delta R$ denote the translational error and rotational error, respectively; 
	$R, t$ is the ground-truth rotation matrix and translation vector; 
	$\tilde{R}, \tilde{t}$ mean the predicted rotation matrix and translation vector;
	$tr(\cdot)$ means the trace of a matrix.
	
	Given the pose errors of all image pairs, we plot the cumulative error distribution curve, whose area at three thresholds $(5^\circ, 10^\circ, 20^\circ)$ are computed as AUC@$(5^\circ, 10^\circ, 20^\circ)$. 
	
	\begin{table}[]
		\centering
		\renewcommand\arraystretch{1.2}
		\caption{\textbf{Indoor pose estimation evaluation} on Scannet dataset. The AUC@(5$^\circ$, 10$^\circ$, 20$^\circ$) is reported. }
		\label{table:1a}
		\resizebox{0.48\textwidth}{!}{
			\begin{tabular}{clccc}
				\toprule[0.3mm]
				\multicolumn{1}{c}{\multirow{2}{*}{\begin{tabular}[c]{@{}c@{}}Local\\ features\end{tabular}}} & \multicolumn{1}{c}{\multirow{2}{*}{Matcher}} & \multicolumn{3}{c}{Pose estimation AUC}          \\ \cline{3-5} 
				\multicolumn{1}{c}{}                                                                          & \multicolumn{1}{c}{}                         & @5$^{\circ}$          & @10$^{\circ}$         & @20$^{\circ}$         \\ \hline
				\multicolumn{5}{c}{\multirow{1}{*}{Detector-based Methods}}
				\\ \hline
				D2-Net \cite{Dusmanu_2019_CVPR}                                                                                       & NN                                  & 5.25           & 14.53          & 27.96          \\
				ContextDesc \cite{luo2019contextdesc}                                                                                   & ratio test \cite{lowe2004distinctive}                             & 6.64           & 15.01          & 25.75          \\  
				\multirow{6}{*}{SuperPoint \cite{DeTone_2018_CVPR_Workshops} }                                                                   & NN                                  & 9.43           & 21.53          & 36.40          \\
				& NN + OANet \cite{zhang2019learning}                                   & 11.76          & 26.90          & 43.85          \\
				& SuperGlue \cite{sarlin2020superglue}                                  & 16.16          & 33.81          & 51.84          \\
				& SGMNet \cite{chen2021learning}                                       & 15.40          & 32.06          & 48.32          \\
				& DenseGAP \cite{kuang2021densegap}                                     & 17.01          & 36.07          & 55.66          \\ 
				& HTMatch \cite{cai2022htmatch}                                     & 15.11          & 31.42          & 48.23          \\\hline
				\multicolumn{5}{c}{\multirow{1}{*}{Detector-free Methods}}
				\\ \hline
				\multirow{6}{*}{------}                                      
				& LoFTR \cite{sun2021loftr}                  & 22.06 & 40.80          & 57.62          \\
				& QuadTree \cite{tang2022quadtree}                  & 24.90 & 44.70          & 61.80          \\
				& MatchFormer \cite{wang2022matchformer}                                        & 24.31          & 43.90 & 61.41 \\
				& ASpanFormer \cite{chen2022aspanformer}                                         & 25.60          & 46.00 & \textbf{63.30} \\
				& DeepMatcher                                         & 25.38          & 44.38 & 60.35  \\
				& DeepMatcher-L                                         & \textbf{27.32}          & \textbf{46.25} & 62.49 \\
				\bottomrule[0.3mm]
				\vspace{-10pt}
		\end{tabular}}
		\vspace{-1em}
		\label{Scannet}
	\end{table}

	\textbf{Results.} 
	As illustrated in \cref{Scannet}, we observe that the detector-free methods achieve superior performance than detector-based methods since the detector struggles to extract repeatable keypoints when handling image pairs with significant viewpoint change. 
	Wherein, DeepMatcher and DeepMatcher-L outperform all cutting-edge detector-based and detector-free methods by a great margin. 
	More specifically, DeepMatcher-L surpasses detector-based method DenseGAP by $(10.31\%, 10.18\%, 6.83\%)$ in terms of AUC@$(5^\circ, 10^\circ, 20^\circ)$, demonstrating the superiority of detector-free structure. 
	Compared with the pioneering method LoFTR, DeepMatcher-L realizes superior performance with the improvement of $(5.26\%, 5.45\%, 4.87\%)$.
	Furthermore, DeepMatcher-L outperforms the state-of-the-art detector-free method ASpanFormer by $(1.72\%, 0.25\%)$ in terms of AUC@$(5^\circ, 10^\circ)$, proving the deep Transformer architecture is essential to extract more human-intuitive and easier-to-match features.
	Additionally, DeepMatcher-L only consumes $77.65\%$ GFLOPs and achieves $26.95\%$ inference speed boost compared with ASpanFormer, as demonstrated in \cref{time_flops}.
	
	\subsection{Outdoor Pose Estimation}
	Outdoor pose estimation remains a challenging task owing to the intricate 3D geometry, extreme illumination and viewpoint changes. 
	To demonstrate the efficacy of DeepMatcher in overcoming these obstacles, an outdoor pose estimation experiment is conducted. 
	
	\textbf{Dataset.} 
	We utilize MegaDepth \cite{li2018megadepth} to conduct the outdoor pose estimation experiment.
	MegaDepth contains 1M internet images from $196$ different scenes.
	These images come from photo-tourism and contain challenging conditions, including large viewpoint and illumination variations.
	Following \cite{sun2021loftr, tyszkiewicz2020disk}, we select 100 image pairs each scene for training and 1500 image pairs for testing. 
	Images are resized such that their longer dimensions are equal to 840.
	
	\textbf{Evaluation Protocol.}
	We use the same evaluation metrics AUC@(5$^\circ$, 10$^\circ$, 20$^\circ$) as the indoor pose estimation task.
	
	\begin{table}[] \small
		\centering
		\renewcommand\arraystretch{1.2}
		\caption{\textbf{Outdoor pose estimation evaluation} on MegaDepth dataset. 
			The AUC@(5$^\circ$, 10$^\circ$, 20$^\circ$) is reported.}
		\label{table:2a}
		\resizebox{0.46\textwidth}{!}{
			\begin{tabular}{clccc}
				\toprule[1pt]
				\multicolumn{1}{c}{\multirow{2}{*}{\begin{tabular}[c]{@{}c@{}}Local\\ features\end{tabular}}} & \multicolumn{1}{c}{\multirow{2}{*}{Matcher}} & \multicolumn{3}{c}{Pose estimation AUC}          \\ \cline{3-5} 
				\multicolumn{1}{c}{}                                                                          & \multicolumn{1}{c}{}                         & @5$^{\circ}$          & @10$^{\circ}$         & @20$^{\circ}$         \\ \hline
				\multicolumn{5}{c}{\multirow{1}{*}{Detector-based Methods}}
				\\ \hline
				\multirow{3}{*}{SuperPoint\cite{DeTone_2018_CVPR_Workshops} }                                                                   & SuperGlue \cite{sarlin2020superglue}                                       & 42.18          & 61.16          & 75.96          \\
				& DenseGAP \cite{kuang2021densegap}                                     & 41.17          & 56.87          & 70.22          \\ 
				& ClusterGNN \cite{shi2022clustergnn}                                     & 44.19          & 58.54          & 70.33          \\ \hline
				\multicolumn{5}{c}{\multirow{1}{*}{Detector-free Methods}}
				\\ \hline
				\multirow{8}{*}{------}    
				& DRC-Net \cite{li2020dual}                                      & 27.01          & 42.96          & 58.31          \\
				& LoFTR \cite{sun2021loftr}                                      & 52.80 & 69.19 & 81.18          \\
				& QuadTree \cite{tang2022quadtree}                                       & 54.60          & 70.50          & 82.20
				\\ 
				& TopicFM \cite{truong2022topicfm}                                       & 54.10          & 70.10          & 81.60
				\\ 
				& MatchFormer \cite{wang2022matchformer}                                       & 52.91          & 69.74          & 82.00
				\\ 
				& ASpanFormer \cite{chen2022aspanformer}                                         & 55.30          & 71.50 & 83.10 \\
				& DeepMatcher                                         & 55.71          & 72.25 & 83.49 \\
				& DeepMatcher-L                                         & \textbf{56.98}          & \textbf{73.11}         & \textbf{84.15} \\
				\bottomrule[1pt]
		\end{tabular}}
		\vspace{-2pt}
		\label{MegaDepth}
	\end{table}
	\textbf{Results.} As shown in \cref{MegaDepth}, we can observe that DeepMatcher families surpass other methods in all evaluation metrics. 
	Specifically, DeepMatcher-L noticeably outperforms the cutting-edge detector-based method ClusterGNN by $(12.79\%, 14.57\%, 13.82\%)$ in AUC@$(5^\circ, 10^\circ, 20^\circ)$ since the detector struggles to extract repeatable keypoints in image pairs with extreme viewpoint change. 
	Besides, compared with the baseline approach LoFTR, DeepMatcher-L achieves superior performance with the improvement of $(4.18\%, 3.92\%, 2.97\%)$.
	Moreover, DeepMatcher-L also surpasses the state-of-the-art detector-free method ASpanFormer by $(1.68\%, 1.61\%, 1.05\%)$, further validating the rationality of the deep Transformer structures.
	
	\subsection{Image Matching}
	As a fundamental visual task, image matching plays an important role in several applications.
	Therefore, we conduct an image matching experiment to validate the performance of DeepMatcher.
	
	\textbf{Dataset.} 
	We conduct homography estimation experiments on the HPatches dataset \cite{balntas2017hpatches}. 
	Following \cite{Dusmanu_2019_CVPR}, we select 108 sequences from HPatches.
	Each sequence consists of a ground-truth homography matrix and 6 images of progressively larger illumination (52 sequences with illumination changes) or viewpoint changes (56 sequences with viewpoint changes).
	
	\textbf{Evaluation Protocol.}
	We adopt the generally employed mean matching accuracy (MMA) as metric, i.e., the average proportion of correct correspondences per image pair \cite{Dusmanu_2019_CVPR}. 
	Specifically, the keypoints from the $i$-th query image are projected into the reference image by using the provided homography matrix $H_{i}$. 
	Then, the matches with reprojection errors that are lower than a predefined threshold $t$ are deemed correct. 
	Finally, we compute the average percentage of correct matches across all image pairs and define MMA as: 
	\begin{equation}
		MMA(t) =\frac{1}{HP}\sum_{i=1}^{HP}(		\frac{\sum_{j=1}^{N^{f}}\mathbbm{1}(t-||H_{i}(P_{A,i,j}^{f}) - P_{B,i,j}^{f}||_{2})}{N^{f}}),
	\end{equation}
	where $HP$ means the number of image pairs in HPatches;
	$N^{f}$ means the number of predicted matches;
	$\mathbbm{1} (\cdot)$ is a binary indicator function whose output is 1 for non-negative value and 0 otherwise;
	$t$ is the threshold of reprojection error, varying from 1 to 10 pixels;
	$H_{i}(\cdot)$ means warping the keypoints in the $i$-th query image to reference image by ground-truth homography matrix;
	$(P_{A,i,j}^{f}, P_{B,i,j}^{f})$ means the pixel coordinates of the $j$-th match in the $i$-th image pair.
	
	\begin{figure}
		\centering
		\includegraphics[width=1.0\hsize]{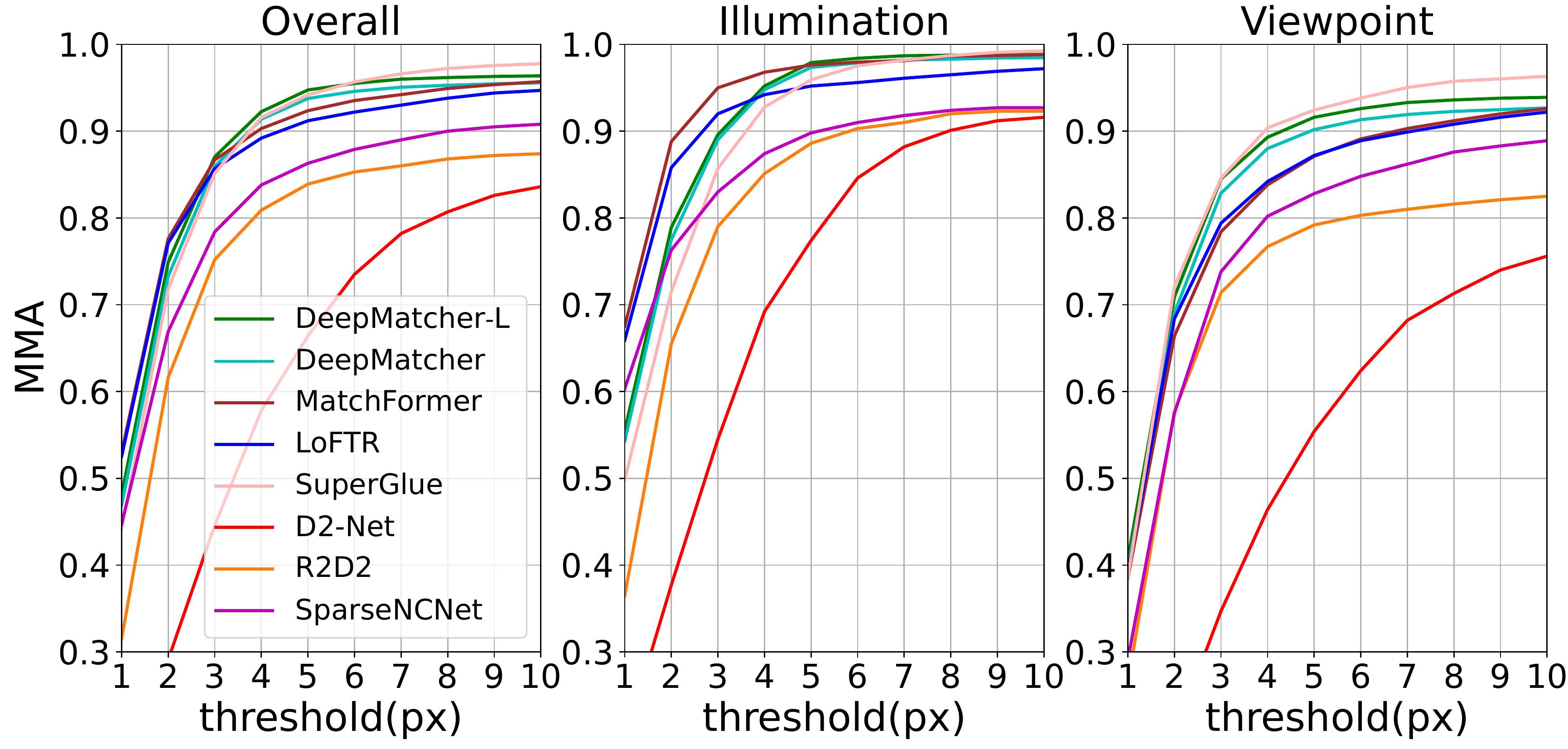}
		\caption{\textbf{Image matching evaluation} on HPatches dataset.}
		\label{MMA}
		\vspace{-10pt}
	\end{figure}
	\begin{figure}[]
		\centering
		\includegraphics[width=0.99\hsize]{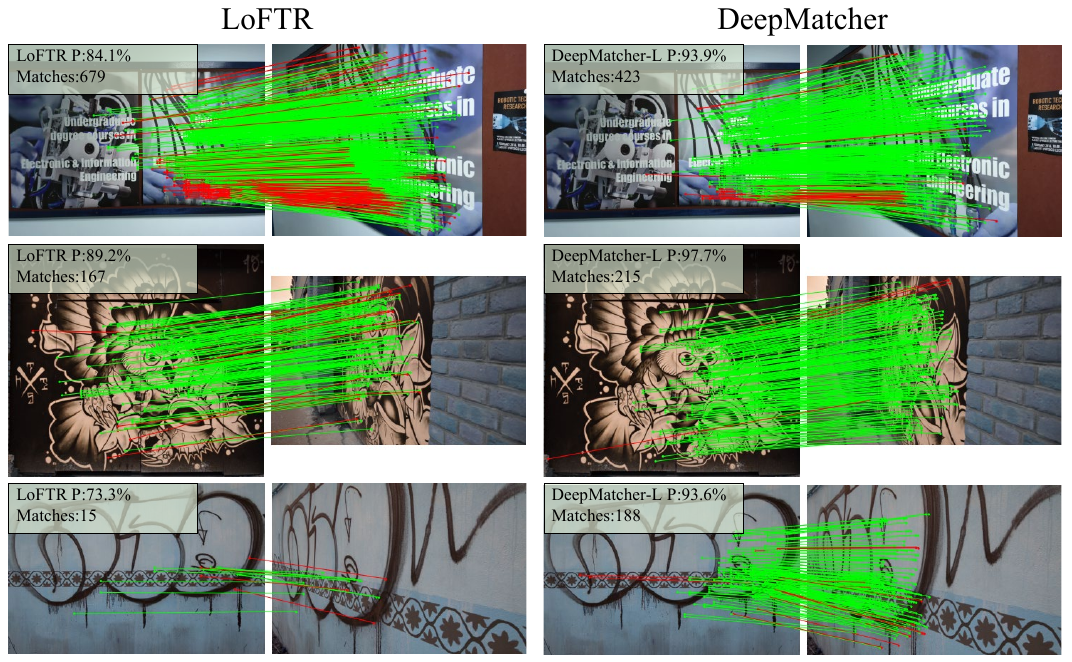}
		\caption{\textbf{Visualization of the predicted matches}.
			The mismatches, whose reprojection errors are larger than 5px, are colored red.}
		\label{large_viewpoint}
	\end{figure}
	
	\textbf{Results.}
	As illustrated in \cref{MMA}, we can observe that DeepMatcher families achieve superior performance than detector-free methods (i.e. MatchFormer, LoFTR, SparseNCNet).
	Under varying illumination conditions, DeepMatcher yields inferior performance at low thresholds, while achieving outstanding performance when the threshold is larger than $5$.
	Moreover, when handling image pairs with viewpoint changes, DeepMatcher exhibits extremely superior robotness compared with other detector-free methods.
	As shown in \cref{large_viewpoint}, we select three pairs of image pairs from HPatches dataset and visualize the matches predicted by LoFTR and DeepMatcher-L to further validate the robustness of DeepMatcher-L to viewpoint variations.
	
	\begin{table}[]\huge
		\centering
		\renewcommand\arraystretch{1.40}
		\caption{\textbf{Homography estimation evaluation} on HPatches dataset.}
		\resizebox{0.48\textwidth}{!}{
			\begin{tabular}{llccc}
				\toprule[2pt]
				\multicolumn{1}{c}{\multirow{2}{*}{\begin{tabular}[c]{@{}c@{}}Local\\ features\end{tabular}}} & \multicolumn{1}{c}{\multirow{2}{*}{Matcher}} & \multicolumn{1}{c}{Overall} & Illumination   & Viewpoint      \\ \cline{3-5} 
				\multicolumn{1}{c}{}                                                                          & \multicolumn{1}{c}{}                         & \multicolumn{3}{c}{CCM ($\varepsilon$\textless 1/3/5 pixels)}              \\ \hline
				\multicolumn{5}{c}{\multirow{1}{*}{Detector-based Methods}}
				\\ \hline
				D2-Net \cite{Dusmanu_2019_CVPR}                                                                                        & NN                                           & 0.38/0.71/0.82              & 0.66/\textbf{0.95}/\textbf{0.98} & 0.12/0.49/0.67 \\
				R2D2 \cite{revaud2019r2d2}                                                                                          & NN                                           & 0.47/0.77/0.82              & 0.63/0.93/\textbf{0.98} & 0.32/0.64/0.70 \\
				ASLFeat \cite{luo2020aslfeat}                                                                                       & NN                                           & 0.48/0.81/0.88              & 0.62/0.94/\textbf{0.98} & 0.34/0.69/0.78 \\ 
				\multirow{4}{*}{SuperPoint \cite{DeTone_2018_CVPR_Workshops} }
				& NN                                           &0.46/0.78/0.85              & 0.57/0.92/0.97 & 0.35/0.65/0.74 \\
				& SuperGlue \cite{sarlin2020superglue}                                   & 0.51/0.82/0.89              & 0.60/0.92/\textbf{0.98} & 0.42/0.71/0.81 \\
				& SGMNet \cite{chen2021learning}                                         & 0.52/\textbf{0.85}/\textbf{0.91}              & 0.59/0.94/\textbf{0.98} & \textbf{0.46/0.74/0.84} \\ 
				& ClusterGNN \cite{shi2022clustergnn}                                         & 0.52/0.84/0.90              & 0.61/0.93/\textbf{0.98} & 0.44/\textbf{0.74}/0.81 \\ 
				\hline
				\multicolumn{5}{c}{\multirow{1}{*}{Detector-free Methods}}
				\\ \hline
				\multirow{6}{*}{------}                                               
				& SparseNCNet \cite{rocco2020efficient}                                 & 0.36/0.65/0.76              & 0.62/0.92/0.97 & 0.13/0.40/0.58 \\
				& Patch2Pix \cite{zhou2021patch2pix}                                   & 0.50/0.79/0.87              & 0.71/\textbf{0.95}/\textbf{0.98} & 0.30/0.64/0.76 \\
				& LoFTR \cite{sun2021loftr}                                         & \textbf{0.55}/0.81/0.86              & 0.74/\textbf{0.95/0.98} & 0.38/0.69/0.76 \\
				& MatchFormer \cite{wang2022matchformer}                                 & \textbf{0.55}/0.81/0.87              & \textbf{0.75}/\textbf{0.95}/\textbf{0.98} & 0.37/0.68/0.78 \\
				& DeepMatcher                                         & 0.50/0.81/0.90             & 0.62/0.93/\textbf{0.98} & 0.38/0.70/0.81 \\ 
				& DeepMatcher-L                                         & 0.51/0.83/\textbf{0.91}           & 0.64/0.94/\textbf{0.98}  & 0.39/0.72/\textbf{0.84} \\ \bottomrule[2pt]
		\end{tabular}}
		\label{homo_experiment}
		\vspace{-10pt}
	\end{table}
	
	\subsection{Homography Estimation}
	Since the distribution and number of matches are essential to estimate reliable geometry relationship between image pairs, we conduct a homography estimation experiment to comprehensively evaluate the performance of DeepMatcher.
	
	\textbf{Dataset.} 
	We assess DeepMatcher on HPatches dataset, which is widely used for homography estimation task.
	\begin{figure*}
		\centering
		\includegraphics[width=1.00\hsize]{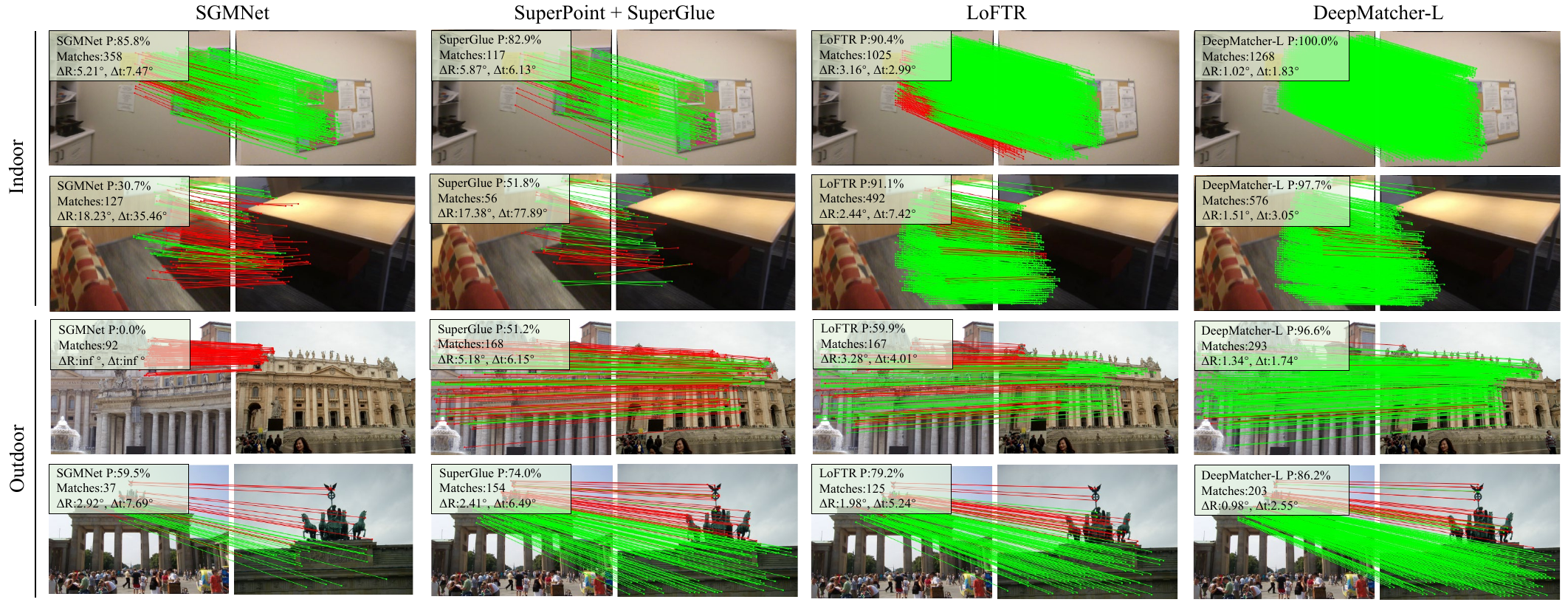}
		\caption{\textbf{Visualization of the predicted matches.}
			The matches are colored by their reprojection errors (green indicates correct matches, and red indicates mismatches).
			We set the error threshold to $10$ and $15$ pixels for indoor and outdoor scenes.} 
		\label{final_matches}
	\end{figure*}
	
	\textbf{Evaluation Protocol.}
	Following the corner correctness metric (CCM) utilized in \cite{zhou2021patch2pix}, we report the percentage of image pairs with average corner errors $\varepsilon$ smaller than 1/3/5 pixels. 
	Specifically, based on the predicted dense matches, we use OPENCV to calculate the homography matrix $\tilde{H}_{i}$ for the $i$-th image pair. 
	Subsequently, four corners in the query image are projected into the reference image by using the ground-truth homography matrix $H_{i}$ and the predicted homography matrix $\tilde{H}_{i}$, respectively. 
	Ultimately, we calculate average reprojection error as corner error $\varepsilon_{i}$, thereby obtaining the CCM:
	\begin{equation}
		\begin{aligned}
			\varepsilon_{i} = & \frac{\sum_{p\in P_{co}}||H_{i}(p) - \tilde{H}_{i}(p)||_{2}}{4}, \\[2pt]
			CC&M(t) = \frac{\sum_{i=1}^{HP}\mathbbm{1}(t-\varepsilon_{i})}{HP},
		\end{aligned}
	\end{equation}
	where $P_{co} = \{(0,0), (W_{o}-1,0), (0,H_{o}-1), (W_{o}-1, H_{o}-1)\}$ means the four corners coordinates of the query image;
	$H_{i}(\cdot)$ and $\tilde{H}_{i}(\cdot)$ mean warping the corners in the $i$-th query image to reference image by ground-truth homography matrix and predicted homography matrix, respectively;
	$t \in \{1,3,5\}$ means the predefined threshold.
	
	\textbf{Results.}
	As shown in \cref{homo_experiment}, DeepMatcher achieves the best performance among the detector-free methods under extreme viewpoint changes.
	Specifically, DeepMatcher outperforms LoFTR and MatchFormer with the improvement of $(1\%, 5\%)$ and $(2\%, 3\%)$ when thresholds are set to $3, 5$ pixels, repsectively.
	Furthermore, DeepMatcher-L surpasses LoFTR and MatchFormer with the improvement of $(1\%, 3\%, 8\%)$ and $(2\%, 4\%, 6\%)$.
	Besides, the detector-based methods are more robust to viewpoint variations, while the detector-free methods realize superior performance when handling image pairs with extreme illumination changes.
	In comparison, DeepMatcher strikes a decent balance when handling image pairs with various viewpoint and illumination changes.

	\begin{figure*}[htbp]
		\centering
		\includegraphics[width=1.0\hsize]{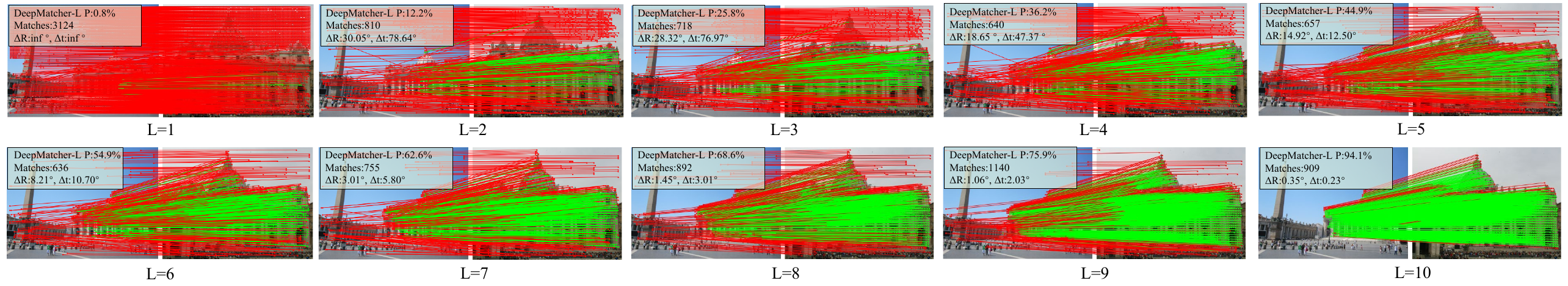}
		\caption{The predicted matches of DeepMatcher-L after each SlimFormer. The threshold $\lambda$ of the soft assignment matrix is set to $0$.} 
		\label{interleave_matches}
		\vspace{-10pt}
	\end{figure*}

	\subsection{Understanding DeepMatcher}
	
	\textbf{Qualitative Results Visualization.}
	To further exhibit the capability of DeepMatcher to handle image pairs with extreme appearance settings, e.g., sparse texture, motion blur, large viewpoint and illumination changes, we visualize the matches predicted by SGMNet, SuperGlue, LoFTR, and DeepMatcher-L.
	As shown in \cref{final_matches}, we can observe that DeepMatcher-L achieves dense and accurate matching performance.
	
	\textbf{Visual Descriptors Enhancement Efficacy Analysis.}
	To validate the effectiveness of performing $L$ times of SlimFormer for feature enhancment, we visualize the matching results of DeepMatcher after each SlimFormer.
	As illustrated in \cref{interleave_matches}, we can observe that the matching precision is promoted consistently, demonstrating interleaving SlimFormer can effectively integrate intra-/inter-image information, hence extracting easier-to-match features.
	
	\begin{table}[] \huge
		\centering
		\renewcommand\arraystretch{1.2}
		\caption{\textbf{Efficiency analysis.} Several approaches are compared in terms of parameters (MB), GFLOPs, and runtime (s). We also record the computational complexity (TC) of the attention layer.
			$N$ denotes the pixle token number, $C$ denotes the feature dimension, $k$ denotes selected token number, $r$ denotes down-scale ratio, $w$ denotes the sample number.}
		\resizebox{0.495\textwidth}{!}{
			\begin{tabular}{l|cccc}
				\toprule[0.9pt]
				\multicolumn{1}{c|}{Methods} & Params & GFLOPs & Runtime & TC \\ \midrule
				LoFTR \cite{sun2021loftr}                        & 11.06       & 328.67 & \textbf{0.079} & $\mathcal{O}(NC^{2})$      \\
				QuadTree \cite{tang2022quadtree}               & 13.21       & 382.01 & 0.152 & $\mathcal{O}(kNC)$      \\
				MatchFormer \cite{wang2022matchformer}                  & 22.37       & 396.95 & 0.357 & $\mathcal{O}(N^{2}C / r)$      \\
				AspanFormer \cite{chen2022aspanformer}                 & 15.05       & 391.38 & 0.141 & $\mathcal{O}((N/r)^{2}C+NwC)$      \\
				DeepMatcher                  & \textbf{10.96}            & \textbf{255.99}       & 0.089 & $\mathcal{O}(NC)$           \\
				DeepMatcher-L                & 15.51       & 303.90 & 0.103 & $\mathcal{O}(NC)$     \\ \bottomrule[0.9pt]
			\end{tabular}
		}
		\vspace{-2pt}
		\label{time_flops}
	\end{table}
	\textbf{Efficiency Analysis.}
	To validate the efficiency of DeepMatcher, we compare several cutting-edge detector-free methods in terms of parameters, flops, and inference speed to determine their computational cost and storage consumption.
	We resize the input images to $640 \times 480$ and conduct all experiments on a single NVIDIA TITAN RTX GPU.
	When counting runtime, we run the test code $500$ times and report the average time to eliminate occasionality.
	Notably, we only compare DeepMatcher families with other detector-free methods.
	As shown in \cref{time_flops}, we can observe that DeepMatcher families realize competitive inference speed with less GFLOPs since SlimFormer leverages element-wise product to model relevance among all keypoints.
	Compared with the baseline LoFTR, DeepMatcher and DeepMatcher-L only consume $(77.89\%, 92.46\%)$ GFLOPs.
	Moreover, compared with cutting-edge detector-free methods QuadTree, MatchFormer and AspanFormer, DeepMatcher-L exhibits more efficient matching performance with $(20.45\%, 23.44\%, 22.35\%)$ less GFLOPs and $(32.24\%, 71.15\%, 26.95\%)$ inference speed boost.
	
	Furthermore, we also record the dominant computational complexity of the attention layers in various methods.
	As shown in \cref{time_flops}, DeepMatcher achieves the minimum computational complexity of the attention layer.
	Specifically, compared with the baseline LoFTR, DeepMatcher reduces the dominant computational complexity from $\mathcal{O}(NC^{2})$ to $\mathcal{O}(NC)$.

	\begin{figure}
		\centering
		\includegraphics[width=0.95\hsize]{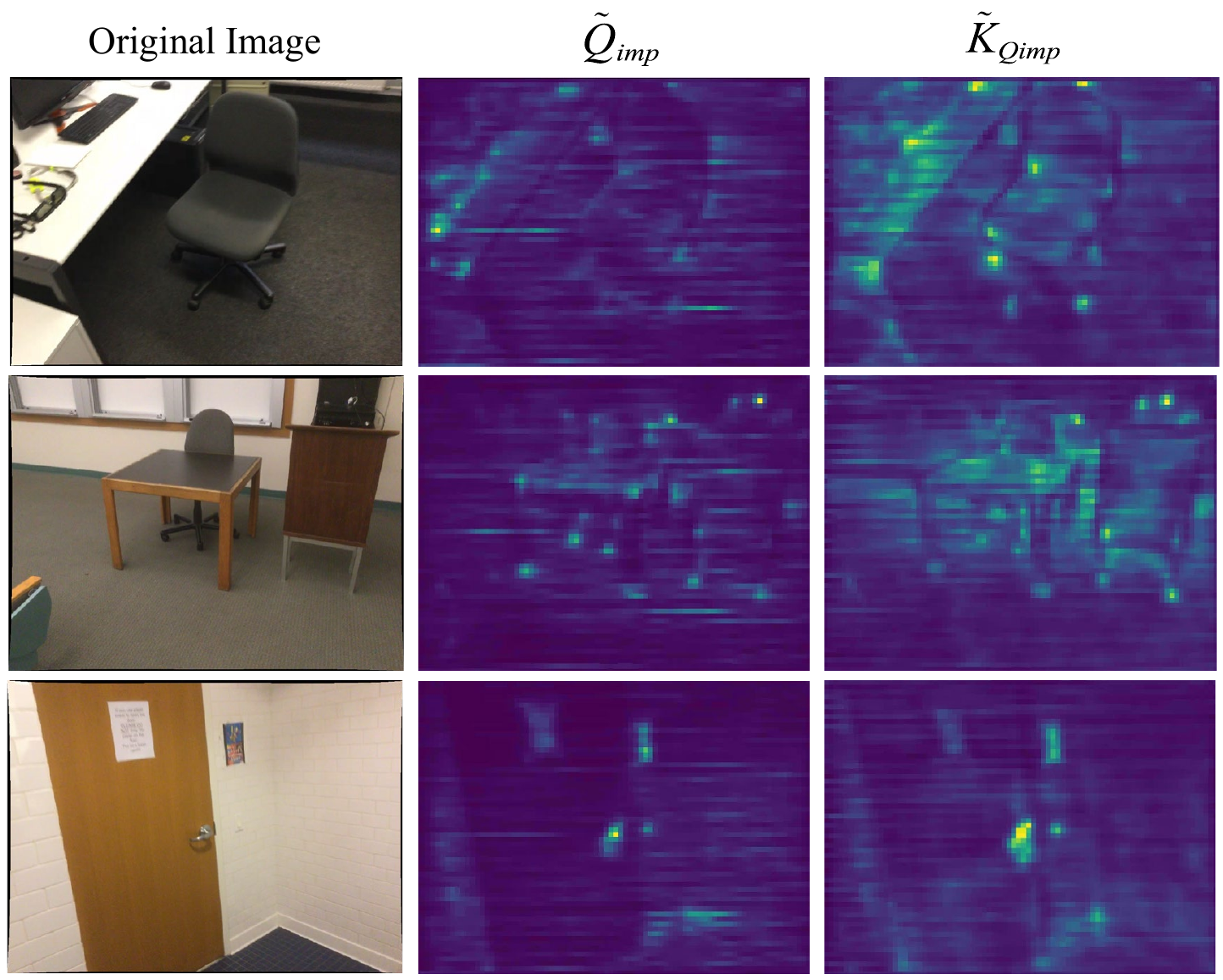}
		\caption{\textbf{Visualization of weights in SlimFormer structure}. SlimFormer emphasises keypoints at object boundaries to incorporate global context.}
		\label{Weight_visual}
		\vspace{-1em}
	\end{figure}
	\begin{figure}[t]
		\begin{minipage}{0.49\linewidth}
			\includegraphics[width=1.0\hsize]{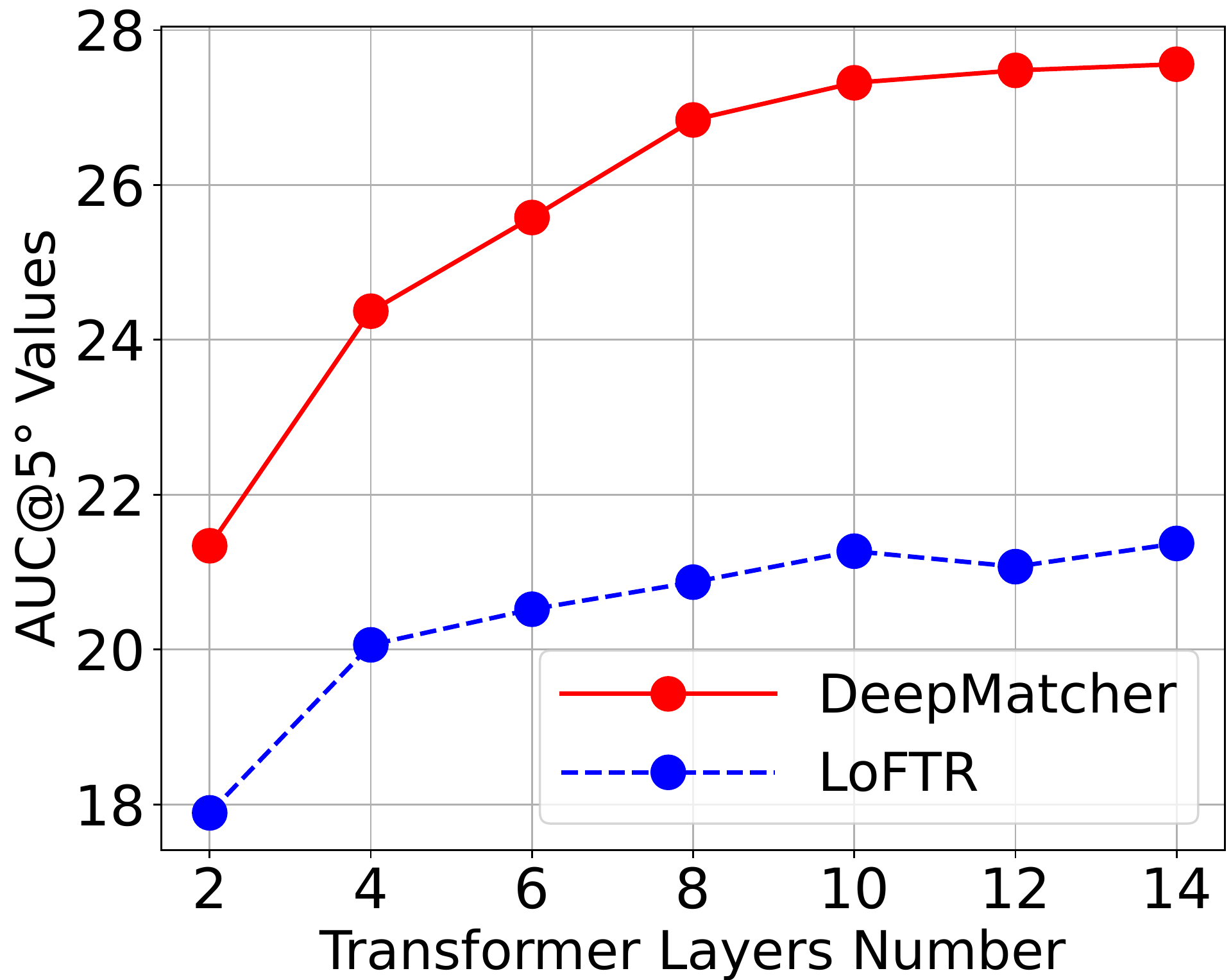}
		\end{minipage}
		\begin{minipage}{0.49\linewidth}
			\includegraphics[width=1.0\hsize]{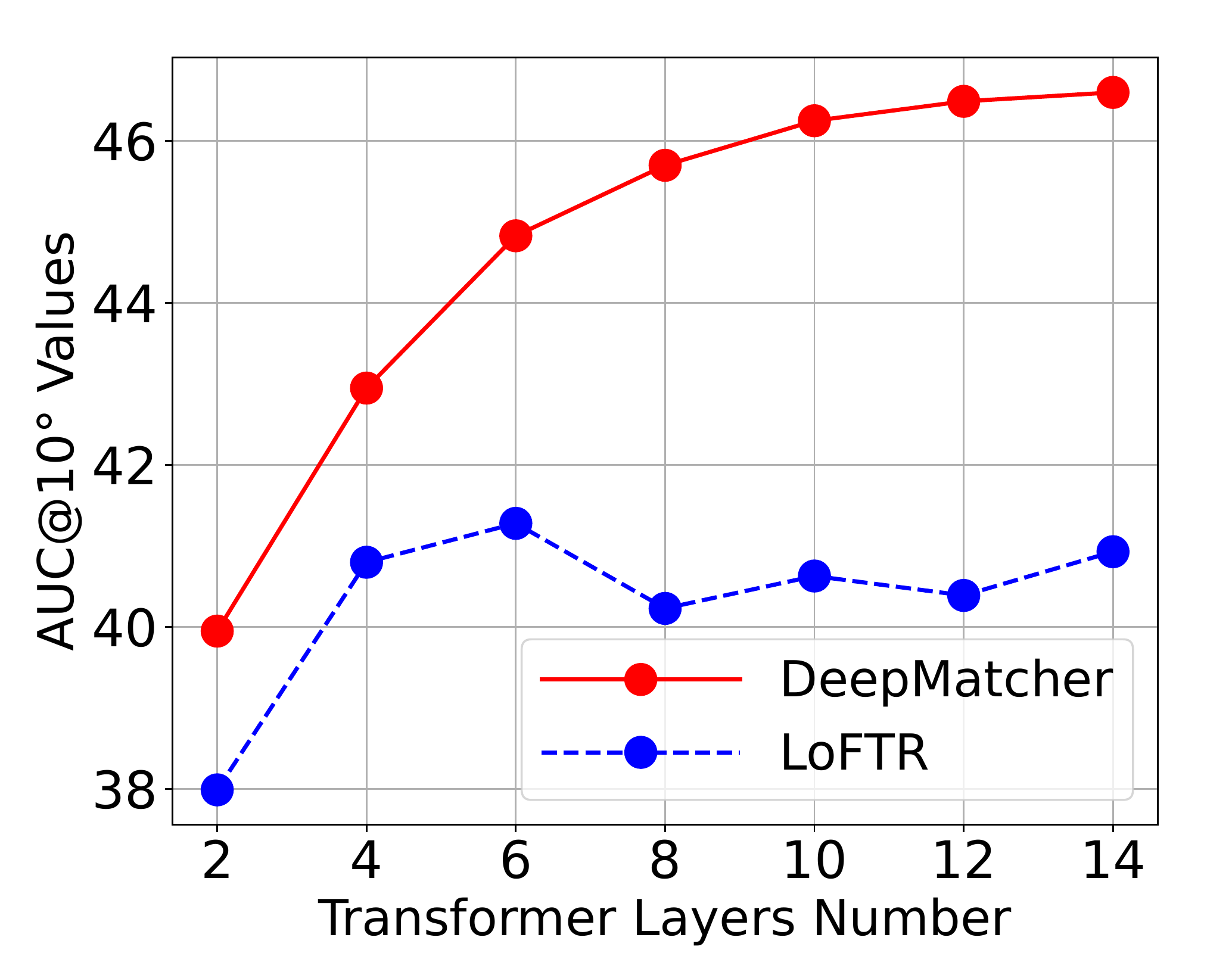}
		\end{minipage}
		\begin{minipage}{0.49\linewidth}
			\includegraphics[width=1.0\hsize]{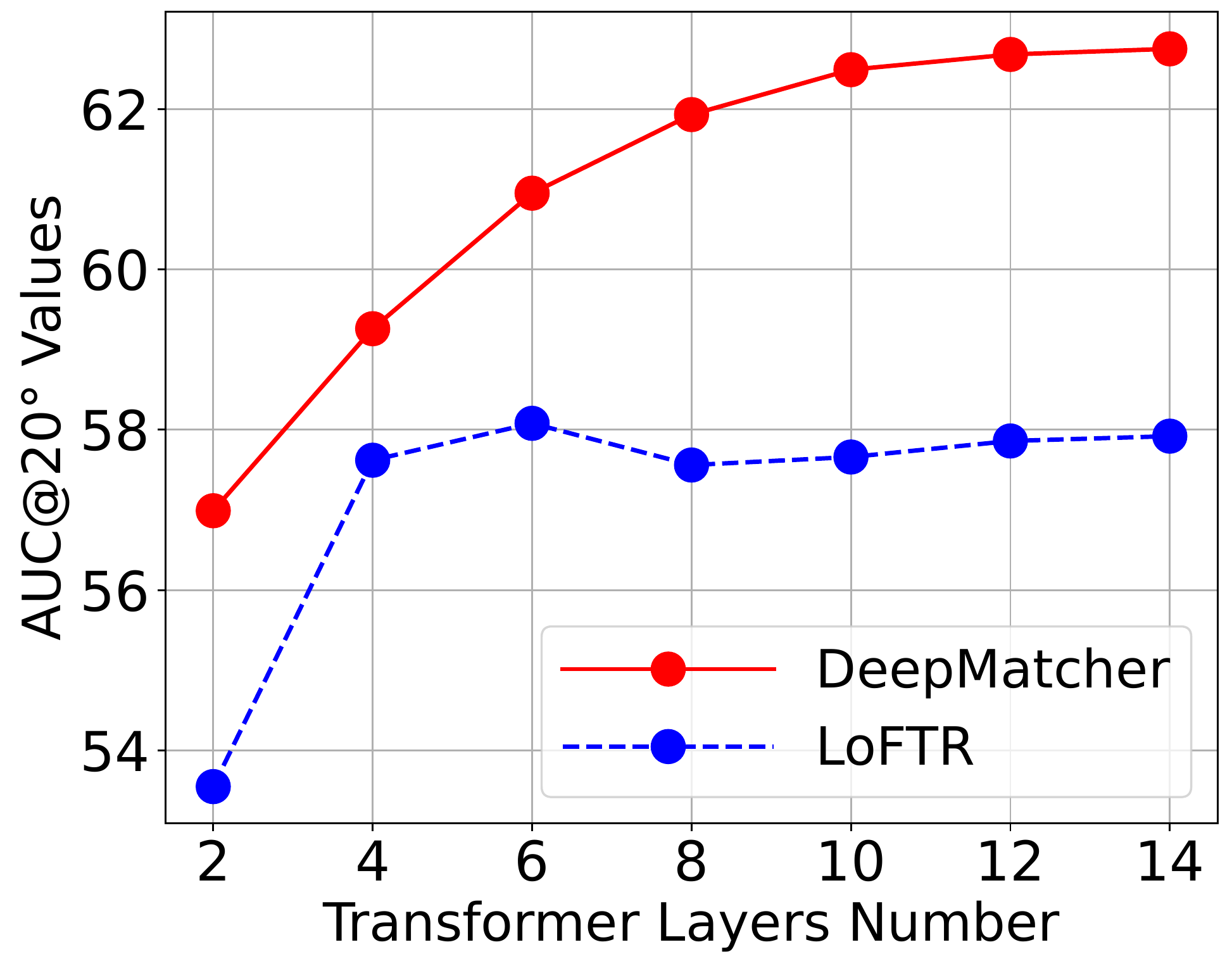}
		\end{minipage}
		\begin{minipage}{0.49\linewidth}
			\includegraphics[width=1.0\hsize]{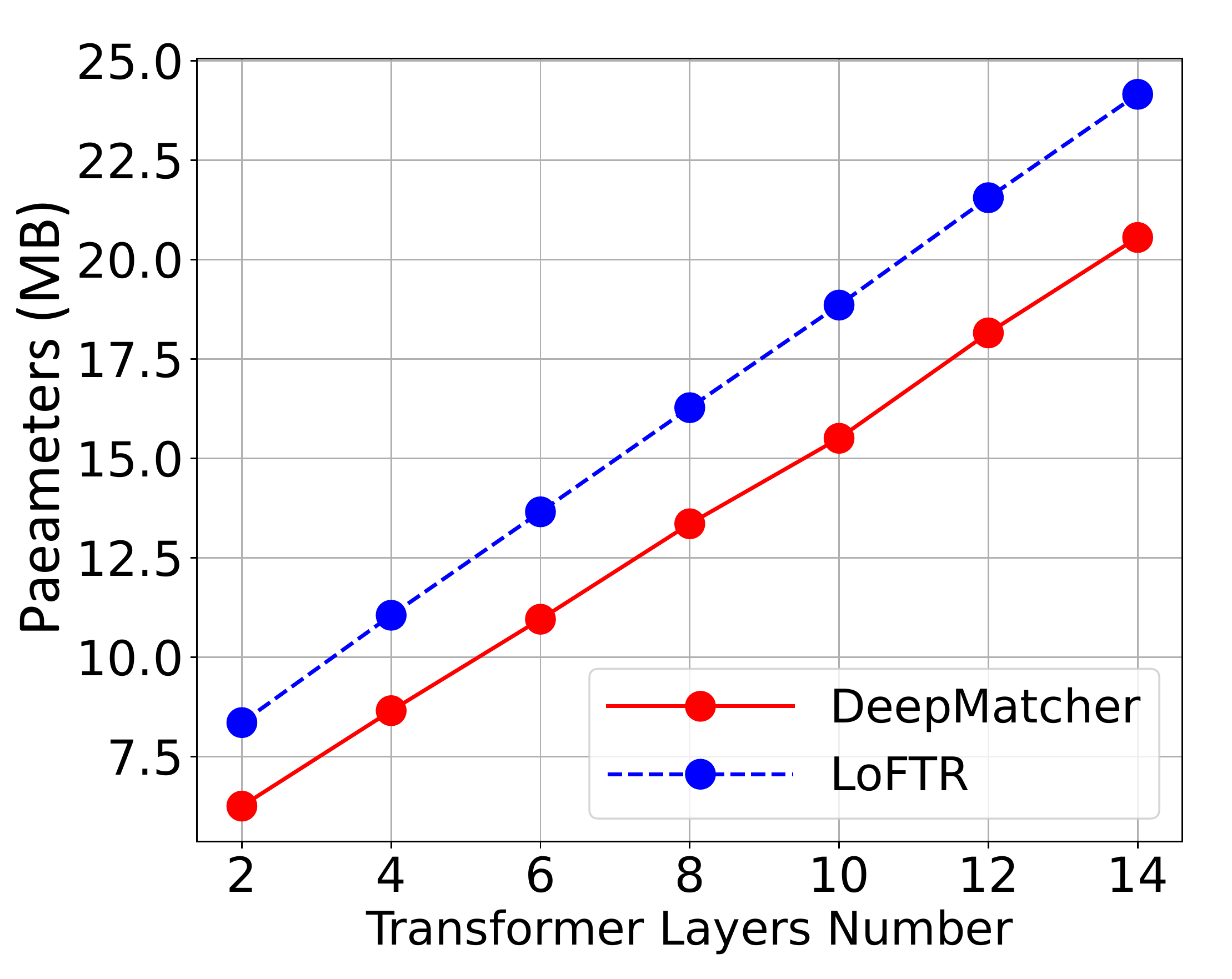}
		\end{minipage}
		
		\caption{\textbf{The AUC values and parameters} of DeepMatcher and LoFTR with the Transformer layers increasing.}
		\label{AUC_Curve}
		\vspace{-1.5em}
	\end{figure}
	\textbf{Weight Analysis.}
	To explore which keypoints SlimFormer pays attention to when extracting global vectors, we visualize the weight $\Tilde{Q}_{imp}$ and $\Tilde{K}_{Qimp}$ in \cref{Qweight} and \cref{KQweight}, respectively.
	As shown in \cref{Weight_visual}, we can observe that SlimFormer primarily pays attention to the prominent keypoints at object boundaries that involve tremendous visual and geometry information.
	Consequently, SlimFormer exhibits puissant capability to aggregate global context information effectively.
	
	\textbf{Deep Transformer Architecture Analysis.}
	To verify the opinion that deep Transformer architecture is essential to extract more human-intuitive and easier-to-match features, we record the indoor pose estimation precision with the number of Transformer layers increasing.
	As shown in \cref{AUC_Curve}, since SlimFormer leverages vector-based attention for robust long-range global context aggregation and utilizes layer-scale strategy and relative position encoding to enhance the representation of
	keypoints, the AUC values of DeepMatcher are promoted consistently with the Transformer layers going deep, while the accuracy of LoFTR is constantly fluctuating.
	Besides, the matching performance of DeepMatcher is significantly superior to LoFTR.
	Moreover, the parameters of DeepMatcher and LoFTR increase linearly with the number of Transformer layers, while DeepMatcher occupies fewer parameters.

	\subsection{Ablation Study}
	\textbf{Effect of the Proposed Modules.}
	To thoroughly validate the rationality of each module, we conduct indoor pose estimation experiments using different variants of DeepMatcher.
	As illustrated in \cref{ablation_experi}, we can observe that all components contribute to the outstanding performance of DeepMatcher.
	
	(i), (ii) Using only self- and cross-SlimFormer layers leads to a severe decrease in matching performance, demonstrating interleaving the self- and cross-SlimFormer layers can effectively integrate intra-/inter-image message.
	(iii) Removing the Feature Transition Module results in a much lower accuracy $(-0.66\%, -0.47\%, -0.33\%)$, proving the effectiveness of ensuring smooth transition between feature extractor and SlimFormer in terms of context ranges.
	(iv) Removing Relative Position Encoding spawns a large drop in pose estimation accuracy $(-1.36\%, -1.67\%, -1.38\%)$, proving the relative position information is crucial to distinguish similar features.
	(v) Removing the Fine Matches Module results in lower AUC values $(-8.34\%, -12.56\%, -14.86\%)$, indicating the effectiveness of refining coarse matches.
	
	\textbf{Effect of the Learnable Scale Factor $\xi$.}
	To validate that layer-scale strategy can simulate the human behaviour that humans can acquire different matching cues each time they scan an image pair to further improve matching performance, we remove the learnable scale factor $\xi$ and conduct an ablation experiment. 
	\begin{table}[t] \large
		\centering
		\renewcommand\arraystretch{1.1}
		\caption{\textbf{Ablation study with different variants of DeepMatcher} on Scannet dataset.}
		\label{struct_ablation}
		\resizebox{0.49\textwidth}{!}{ 
			\begin{tabular}{lccc}
				\toprule[1.2pt]
				\multicolumn{1}{c}{\multirow{2}{*}{Methods}}           & \multicolumn{3}{c}{Pose estimation AUC} \\ \cline{2-4} 
				& @5$^{\circ}$          & @10$^{\circ}$         & @20$^{\circ}$         \\ \hline
				(i) w only self-SlimFormer layers    & 20.38         & 37.71        & 53.89        \\
				(ii) w only cross-SlimFormer layers    & 22.33          & 40.27          & 56.64         \\
				(iii) w/o Feature Transition Module    & 24.72            & 43.91            & 60.02            \\
				(iv) w/o Relative Potition Encoding  & 24.02            & 42.71           & 58.97            \\
				(v) w/o Fine Matches Module    & 17.04            & 31.82            & 45.49            \\
				DeepMatcher full                          & \textbf{25.38}           & \textbf{44.38}            & \textbf{60.35}            \\ \bottomrule[1.2pt]
			\end{tabular}
			\label{ablation_experi}
			\vspace{-10pt}}
	\end{table}
	\begin{table}[]
		\centering
		\caption{\textbf{Ablation study with different learnable scale factors $\xi$} on Scannet dataset.}
		\renewcommand\arraystretch{1.3}
		\resizebox{0.43\textwidth}{!}{%
			\begin{tabular}{lccc}
				\toprule[0.8pt]
				\multicolumn{1}{c}{\multirow{2}{*}{Methods}} & \multicolumn{3}{c}{Pose estimation AUC}                          \\ \cline{2-4} 
				& @5$^{\circ}$             & @10$^{\circ}$            & @20$^{\circ}$            \\ \hline
				w residual scale factor $\xi$                                    &  \textbf{25.38}        & \textbf{44.38}          &  \textbf{60.35}       \\
				w/o residual scale factor $\xi$                                    & 24.50      & 42.91          &  59.32 
				\\ \bottomrule[0.8pt]
		\end{tabular}}
		\label{scalefactor_table}
		\vspace{-1em}
	\end{table}
	As shown in \cref{scalefactor_table}, we can observe that introducing the residual scaling factor leads to superior performance. 
	
	\begin{table}[t]
		\centering
		\caption{\textbf{Ablation study with different position encoding} on Scannet dataset.}
		\renewcommand\arraystretch{1.3}
		\resizebox{0.43\textwidth}{!}{%
			\begin{tabular}{lccc}
				\toprule[0.8pt]
				\multicolumn{1}{c}{\multirow{2}{*}{Methods}} & \multicolumn{3}{c}{Pose estimation AUC}                          \\ \cline{2-4} 
				& @5$^{\circ}$             & @10$^{\circ}$            & @20$^{\circ}$            \\ \hline
				w/o position encoding                                    &  24.02        & 42.71          & 58.97         \\
				w absolute position encoding                                    &  24.52     & 43.14          &  59.51 \\
				w relative position encoding                                    &  \textbf{25.38}        & \textbf{44.38}          & \textbf{60.35} 
				\\ \bottomrule[0.8pt]
		\end{tabular}}
		\label{positon_table}
	\end{table}
	\begin{table}[htbp]
		\centering
		\caption{\textbf{Ablation study with different coarse-to-fine modules} on Scannet dataset.}
		\renewcommand\arraystretch{1.3}
		\resizebox{0.43\textwidth}{!}{%
			\begin{tabular}{lccc}
				\toprule[0.8pt]
				\multicolumn{1}{c}{\multirow{2}{*}{Methods}} & \multicolumn{3}{c}{Pose estimation AUC}                          \\ \cline{2-4} 
				& @5$^{\circ}$             & @10$^{\circ}$            & @20$^{\circ}$            \\ \hline
				Coarse-to-fine Module in LoFTR                                   & 23.74         & 42.26          & 58.86         \\
				Fine Matches Module                                    &  \textbf{25.38}        & \textbf{44.38}          & \textbf{60.35} 
				\\ \bottomrule[0.8pt]
		\end{tabular}}
		\label{finemodule_table}
		\vspace{-1.5em}
	\end{table}
	
	\textbf{Effect of the Relative Position Encoding.}
	Humans leverage the relative position information to establish the connection between objects.
	Therefore, the relative position encoding is more conducive to realize elaborate scene parsing.
	To prove this opinion, we implement an ablation experiment using three structures:
	(i) Removing all position encoding in all SlimFormer.
	(ii) Using absolute position encoding proposed in LoFTR.
	(iii) Using relative position encoding.
	As shown in \cref{positon_table}, we can observe that both absolute and relative position encoding boost AUC values, in which the relative position encoding exhibits more superior performance than absolute position encoding with the improvement of $(0.86\%, 1.24\%, 0.84\%)$.

	\textbf{Effect of the Fine Matches Module.}
	Compared with the coarse-to-fine module used in LoFTR, FMM views the match refinement as a combination of classification and regression tasks.
	To validate the availability of FMM, we conduct an ablation experiment.
	As shown in \cref{finemodule_table}, we can observe that using FMM significantly achieves superior performance with the improvement of $(1.64\%, 2.12\%, 1.49\%)$, proving the rationality of predicting offset and confidence concurrently using a network.
	
	

	\section{Conclusion}
	In this work, we propose DeepMatcher, a deep Transformer-based network for local feature matching. 
	DeepMatcher simulates human behaviors when humans match image pairs, including: (1) Deep SlimFormer layers of the network to aggregate information intra-/inter-images; (2) Layer-scale strategy to assimilate message exchange from each layer adaptively. 
	Besides, relative position encoding is applied to each layer so as to explicitly disclose relative distance information, hence improving the representation of DeepMatcher. 
	We also propose Fine Matches Module to refine the coarse matches, thus generating robust and accurate matches. 
	Extensive experiments demonstrate that DeepMatcher surpasses state-of-the-art approaches on several benchmarks, confirming the superior matching capability of DeepMatcher.
	
	\bibliographystyle{IEEEtran}
	\bibliography{IEEEabrv,bib.bib}
\end{document}